\RequirePackage[hyphens]{url}
\documentclass[11pt]{article} 
\usepackage[margin=2.5cm]{geometry}
\usepackage{setspace}
\usepackage{amsmath,amssymb,amsfonts}
\usepackage{bm}
\DeclareMathOperator*{\argmax}{arg\,max}

\usepackage{natbib}
 \bibpunct[, ]{(}{)}{,}{a}{}{,}

\usepackage[toc,page, title, titletoc]{appendix}
\usepackage{rotating}
\usepackage{fancyvrb}
\usepackage{multirow}
\usepackage{array}
\usepackage{algorithm}
\usepackage{algpseudocode}
\usepackage{setspace}
\usepackage[export]{adjustbox}
\usepackage{anyfontsize}
\usepackage{booktabs}
\usepackage{longtable}
\usepackage[hidelinks]{hyperref}
\usepackage{caption}

\usepackage{fancyhdr}
\begin{document}

\title{Honesty is the Best Policy ... or is it? Demand Management for Probabilistic Service Options\vspace{-4ex}}
\date{\vspace{-4ex}}
\maketitle
\begin{center}
Daniela Sailer$^1$, Robert Klein$^1$, Claudius Steinhardt$^2$ \par
\footnotesize
daniela.sailer@wiwi.uni-augsburg.de, robert.klein@wiwi.uni-augsburg.de, \par
claudius.steinhardt@unibw.de
\end{center}
\begin{center}
\footnotesize
  $^1$ Chair of Analytics \& Optimization, University of Augsburg \par  
  $^2$ Chair of Business Analytics \& Management Science, University of the Bundeswehr Munich
\end{center} 

\begin{abstract}
    To enable well-informed customer choices and redistribute demand across service options, providers can display probabilities regarding predicted service levels at the time of booking, effectively offering probabilistic service options. In many service settings, the underlying truthful probabilities endogenously depend on the system state, rather than being static product attributes controlled by the provider. While customers rely on this information when making decisions, they cannot readily verify its accuracy. Consequently, providers may be tempted to manipulate the displayed probabilities to guide customers’ choices in their favor, creating a novel lever for demand management. In this work, we model the resulting problem as a sequential decision process, define truthful probabilities recursively, and introduce transparency constraints that limit their distortion. Because the exact computation of truthful probabilities is intractable for realistic settings, we develop an adapted reinforcement learning algorithm to approximate them. Furthermore, we propose a sampling-based mixed-integer program to optimize the information communicated by the provider in each state. Our framework enables providers to gauge the benefits of dynamically disclosing probabilistic information at the time of booking and sheds light on incentives for manipulation. Our computational study is motivated by out-of-home delivery, i.e., the delivery to parcel lockers or shops, and reveals a trade-off between individual-level transparency and aggregate customer utilities.
\end{abstract}


\section{Introduction} \label{intro}
Services are fundamentally characterized by the active participation of customers in the value creation process \citep[][Chapter 1]{bordoloi_etal_2023}. From the provider's perspective, this introduces substantial uncertainty regarding the available capacity given that the amount or duration of resources being blocked may also depend on customers' behavior. Consequently, guaranteeing service provision for every advance reservation would require a conservative definition of feasibility based on worst-case assumptions regarding resource availability, which entails high costs in the form of foregone revenue and underutilized capacity \citep{sailer_etal_2024}.

To sidestep strict feasibility requirements at the time of booking, service providers have devised strategies to deal with insufficient capacity ex post at the time of service provision. On the one hand, providers may deny the originally promised service, resorting to backup options or compensation schemes. Examples include widespread overbooking practices in the travel industry, preemptible instances in cloud computing, or the rerouting of parcels in out-of-home delivery networks. On the other hand, providers may still accommodate all customers at the cost of degraded service quality, such as throttled speed at electric vehicle charging stations or standing passengers on public transit. In both cases, customers systematically prefer one service outcome (receiving service, higher quality) over the other.

While these strategies enable providers to efficiently manage uncertainty, they shift most of the associated risks and inconvenience to customers, causing dissatisfaction. To mitigate this, the provider can share information on predicted service levels at the time of booking, thereby unlocking two key advantages: First, communicating predicted congestion or availability supports well-informed customer choices that align with individual preferences and risk tolerance. Second, the information simultaneously serves as a demand-shaping instrument because it induces customers to redistribute themselves across service options, reducing inefficient demand concentration and the need for ex post interventions.

In this work, we provide such information in the form of probabilities, thereby constructing \textit{probabilistic service options}. The probabilities are \textit{displayed} to dynamically arriving \textit{requests} at the time of booking and represent the likelihood of receiving a capacitated \textit{premium outcome} (the higher quality level or not being denied service altogether) at the time of service provision. With the remaining probability, the customer receives the service option's \textit{base outcome}. Essentially, the provider offers an assortment of probabilistic service options to each incoming request. In response, the customer selects one of them (possibly including the alternative not to book), turning the request into an \textit{order}. At specific points in time, service outcomes are \textit{allocated} to orders. We allow for resources to be reusable, i.e., customers make use of the capacity for a (potentially random) duration, and it becomes available again afterward. In total, we consider three sources of uncertainty: request arrival, customer choice, and usage duration.

From a conceptual perspective, these probabilistic service options resemble probabilistic products \citep{fay&xie_2008, goensch_2020}, i.e., an offering that comprises multiple component products, each associated with a probability. Existing research on probabilistic products investigates probabilities as a static design choice, set exogenously by the provider prior to selling. In contrast, the stochasticity of the realized outcome in service settings is not the provider's deliberate choice, but naturally emerges from the uncertain availability of capacity. Such uncertainty may stem from random capacity usage durations, customer no-shows, or orders displacing earlier arrivals if allocation decisions do not follow the first-come-first-served principle. As a result, the truthful probabilities endogenously depend on the state of the system, potentially including the characteristics of the current request, and vary dynamically during the selling horizon.

Unlike traditional supervised learning, simply estimating the probabilities from historical data would lead to biased predictions because the data reflects customers' choices under probabilities displayed in the past. Once the provider communicates newly estimated probabilities, customers' choices adapt in response, and hence the underlying true probabilities change, resulting in a feedback loop. To communicate accurate probabilities to sequentially arriving requests, the provider must forecast demand, anticipate choice behavior of future requests in response to probabilities expected to be displayed in subsequent periods, predict allocation decisions, and account for units already sold along with current utilization. For realistic problem sizes, a vast space of possible states paired with the continuous probability space and customers' expectation to receive probabilities in real time pose significant challenges.

\subsection{Contributions} \label{intro_contributions}
To the best of our knowledge, we are the first to study endogenously probabilistic products, thereby enabling providers to assess the benefits and implications of dynamically communicating probabilistic information. Moreover, because customers cannot readily verify their accuracy, providers may be tempted to manipulate the displayed probabilities to guide customers' choices in their favor. Beyond controlling product availability or prices, this creates a novel, information-based lever for demand management that may call for regulation. In response, we propose transparency constraints that limit the deviation between displayed and truthful probabilities, with full transparency referring to the case where the provider is honest and always reports probabilities equal to the truthful ones. Our work provides methodology to quantify the extent of the provider's incentive for distortion and assess how transparency constraints affect the involved stakeholders.

With our work, we make the following contributions:
\begin{itemize}
    \item \textit{Modeling framework.} We formalize the problem as an infinite-horizon sequential decision process. As key modeling contributions, we formulate a recursive equation to compute truthful, endogenous probabilities under a given policy and introduce transparency constraints. 
    \item \textit{Solution methodology.} As a theoretical insight, we establish that, for certain allocation schemes, there exists exactly one feasible respective solution for the displayed probabilities under full transparency. For realistic problems, the exact computation of truthful probabilities is intractable. In response, we propose an adapted reinforcement learning algorithm to train an approximation. To optimize displayed probabilities, we develop a mixed-integer program that incorporates customer choice behavior through a sampling approach.
    \item \textit{Managerial insights.} Our computational study is motivated by out-of-home delivery (OOHD), i.e., the delivery to parcel lockers or shops instead of the recipient's doorstep \citep{janinhoff_etal_2024}. In practice, providers reroute parcels to alternative facilities if the one selected by the customer is fully occupied, effectively following an ex post strategy to handle insufficient capacity. Given the substantial cost savings associated with OOHD, providers seek to encourage adoption by maximizing customer satisfaction. Our study yields three core insights: First, while truthful probability disclosure already improves aggregate customer satisfaction substantially, demand management yields further gains through targeted adjustments of the displayed probabilities. Notably, marginal gains are diminishing with the permitted amount of probability distortion. Second, the value of information provision and demand management is greatest when customers strongly prefer scarce premium outcomes and for high-density networks that offer numerous suitable alternatives. Third, these aggregate benefits must be weighed against systematic utility gains and losses among customers as well as potential long-term effects on credibility.
\end{itemize}

\subsection{Related Work} \label{intro_related_work}
Our work lies at the intersection of probabilistic selling, revenue management, and information provision. We concisely review the corresponding literature streams, delineating our contributions from existing work.

\subsubsection{Probabilistic Selling} \label{intro_rw_probabilistic_selling}
Existing studies on probabilistic selling exclusively take on a strategic perspective, comparing selling strategies \citep{fay&xie_2008, guo_etal_2023} or investigating product design \citep{chen_etal_2021, zhang_etal_2015}. With the exception of \cite{mitter_etal_2026}, publications predominantly address settings with a single probabilistic product, typically comprising two vertically differentiated component products \citep{chen_etal_2024, guo_etal_2023, ren&huang_2022, zhang_etal_2015}. 

Across the entire literature stream, allocation probabilities are assumed to be static. They are either decided by the seller \citep{mitter_etal_2026} or directly result from the inventory at the start of the selling horizon \citep{ren&huang_2022, zhang_etal_2015}. Such static probabilities are justified if there are no capacity constraints \citep{chen_etal_2021} or a fixed capacity is reserved for the probabilistic product without availability information being revealed over time \citep{yang_etal_2026}.

Taking on a novel operational perspective, we extend probabilistic selling to settings with dynamic, endogenous probabilities for assortments comprising multiple probabilistic products, thereby transforming it into an instrument to improve transparency in the face of inherently uncertain capacity.

\subsubsection{Revenue Management} \label{intro_rw_revenue_mgmt}
Similar to probabilistic products, flexible products in revenue management allow sellers to allocate an associated component product after the sale, but without announcing probabilities beforehand \citep{goensch_2020}. The literature treats such offerings as a deliberate design choice to gain operational flexibility \citep{goensch_etal_2014} and an additional dimension for market segmentation by targeting customers willing to accept uncertainty, typically in exchange for a lower price \citep{gallego&phillips_2004}.

Research in this domain focuses on dynamic decision making, controlling product availability \citep{cheung&simchi-levi_2016, koch_etal_2017} or prices \citep{strauss_etal_2021} over time. The sale of a flexible product is permitted even when some of its component products are guaranteed to be unavailable, as long as at least one feasible allocation exists \citep{petrick_etal_2010}. While allocation typically occurs once, i.e., within \citep{petrick_etal_2012} or at the end of the finite selling horizon, \cite{zhu&topaloglu_2024} also consider periodic allocation decisions.

Transferring the operational perspective of revenue management to probabilistic selling, we employ information provision in the form of allocation probabilities as a novel mechanism to dynamically steer demand instead of availability or prices. 

\subsubsection{Information Provision} \label{intro_rw_provision}
Beyond probabilities, the literature proposes other forms of information that providers may share with customers:

\begin{itemize}
    \item In queuing-based service systems, the current queue length or estimated waiting time help customers decide whether to join the queue \citep{ibrahim_2018}. Going beyond settings where service is executed as soon as possible, our work extends to advance reservations, incorporating a deliberate time lag between booking and service provision commonly encountered in practice. 
    \item A second branch of literature examines messages sent to all or a subset of customers, e.g., regarding remaining inventory. In the majority of publications, messages to all targeted customers are dispatched simultaneously at a single point in time rather than sequentially \citep{allon&bassamboo_2011, drakopoulos_etal_2021, cui&shin_2018}. While \cite{simhon&starobinski_2018} dynamically display the total number of unsold seats over a finite booking horizon, our framework features uncertain and reusable capacity. As a result, the available capacity at the time of service provision cannot be determined with certainty in advance in our case because it does not directly follow from the number of units sold, necessitating more nuanced information.
    \item In the context of restaurant meal delivery, \cite{hildebrandt&ulmer_2022} focus on accurately predicting estimated arrival times for a given routing and assignment policy. In comparison, we additionally consider the notion of deliberately distorting information to manage demand.
\end{itemize}

\subsection{Organization of the Paper} \label{intro_org}
The remainder of this paper is structured as follows: In Section~\ref{model}, we explain our modeling framework. Section~\ref{solution_approach} is dedicated to our solution methodology. We present the results of our computational study in Section~\ref{comp_study} and conclude the paper with future research directions in Section~\ref{conclusion}.

\section{Modeling Framework} \label{model}
In this section, we define our problem (Section~\ref{problem_description}) and formalize it as a sequential decision process (Section~\ref{SDP}). As a novel modeling component, we propose a recursive equation to compute truthful allocation probabilities (Section~\ref{SDP_ground_truth}). Based on them, we formulate transparency constraints, resulting in a constrained variant of the sequential decision process (Section~\ref{sdp_transparency}). Appendix~\ref{overview_notation} provides an overview on the notation.

\subsection{Problem Definition} \label{problem_description}
We denote the set of all service outcomes by $\mathcal{I}=\mathcal{I}^-\cup\mathcal{I}^+=\{1,\dots,I\}$, partitioned into base outcomes $\mathcal{I}^-$ and premium outcomes $\mathcal{I}^+$. The capacity $Q_i$ of premium outcomes $i\in\mathcal{I}^+$ is limited, whereas capacity constraints for base outcomes are assumed to be negligible, either because capacity is unrestricted or has been verified as sufficient ex ante. Each probabilistic service option $j\in\mathcal{J}^P$ corresponds to a unique pair $(i_j^{-},i_j^{+})$ of a base outcome $i_j^{-}\in\mathcal{I}^-$ and a premium outcome $i_j^{+}\in\mathcal{I}^+$. Across different application contexts, examples include indoor versus terrace seating at a restaurant, delivery to a backup parcel shop versus the originally selected parcel locker, and slow versus fast electric vehicle charging. Base outcomes may additionally be offered as deterministic options $j\in\mathcal{J}^D$ with $i_j^+=i_j^-=i_j\in\mathcal{I}^-$. The possibility that the customer chooses not to book can also be modeled with an artificial deterministic option $j\in\mathcal{J}^D$. Set $\mathcal{J}=\mathcal{J}^P\cup\mathcal{J}^D$ represents all offered options.

The planning horizon is divided into time periods such that at most one request arrives per period, yielding a set of discrete points in time $t\in\mathcal{T}=\{1,\dots,\infty\}$. Whenever a new request arrives, the provider displays a probability $p_j$ for every probabilistic service option $j\in\mathcal{J^P}$ (e.g., restaurants, delivery locations, charging stations). A customer choosing $j\in\mathcal{J^P}$ expects to receive the premium outcome $i_j^+$ with probability $p_j\in[0,1]$ and the base outcome $i_j^-$ with the complementary probability $1-p_j$. Based on the displayed probabilities, the customer chooses one of the options $j\in\mathcal{J}$, and the request turns into an order.

At designated points in time $t\in\mathcal{T}^A\subset\mathcal{T}$, the provider fulfills orders scheduled for service provision by allocating service outcomes. For each order $o$, the point in time $t_o^A\in\mathcal{T}^A$ at which allocation for $o$ must occur is known and cannot be pre- or postponed. Allocating a premium outcome temporarily reduces the available capacity for a (stochastic) amount of time. Because our analysis focuses on the role of (in)transparency in displayed probabilities as a demand management lever, we assume that the provider does not strategically withhold available capacity during allocation and always allocates the premium outcome to an order if sufficient capacity is available. Otherwise, we would introduce a second source of allocation-related intransparency. The provider seeks to maximize the expected average reward, e.g., profit or customer satisfaction.

\subsection{Sequential Decision Process} \label{SDP}
Leveraging the framework by \cite{powell_2022}, we formalize the basic problem without transparency constraints as a sequential decision process.

\subsubsection{Decision Epoch} \label{sdp_epochs} 
A decision epoch $k=1,\dots,\infty$ occurs whenever a new request arrives, prompting the provider to specify probabilities, or the system reaches a point in time for service provision, necessitating an allocation decision.

\subsubsection{Pre-Decision State} \label{sdp_preds}
The pre-decision state variable $S_k$ comprises four components:
\begin{enumerate}
    \item \textit{Time.} The current point in time is denoted by $t_k$. 
    \item \textit{Request.} We encode the newly arrived request with $r_k$. Beyond the finite point in time $t_{r_k}^A\in\mathcal{T}^A$ for allocation, the provider may observe further request attributes, such as the customer's segment, that correlate with choice behavior or the capacity usage duration. We use $r_k=0$ to model the case of no request arrival.
    \item \textit{Pending orders.} The set $\mathcal{O}_k$ represents all pending orders, i.e., orders that have not been allocated yet. For each pending order $o$, we track the service option $j_o$ chosen by the customer and the point in time $t^A_o\in\mathcal{T}^A$ at which one of the eligible service outcomes $i_{j_o}^-$ and $i_{j_o}^+$ must be allocated. In addition, further information such as the order's arrival time may be necessary (Section~\ref{comp_study}).
    \item \textit{Capacity utilization.} For each premium service outcome $i\in\mathcal{I}^+$, the set $\mathcal{C}_{ik}$ contains all orders that currently occupy its capacity as a result of previous allocation decisions, including information on the amount of blocked capacity and the elapsed usage duration. From this, we can infer the available capacity $q_i^k$ in decision epoch $k$. We summarize the capacity utilization with $\mathcal{C}_k=\{\mathcal{C}_{ik}:i\in\mathcal{I}^+\}$.
\end{enumerate}

In summary, we define the pre-decision state variable as $S_k=(t_k,r_k,\mathcal{O}_k,\mathcal{C}_k)$ for $k>0$ with initial state $S_0$ and model the set of all possible pre-decision states with the pre-decision state space~$\mathcal{S}$.

\subsubsection{Decision} \label{sdp_decision} 
We generally denote decisions by $X_k\in\mathcal{X}(S_k)$ with the decision space $\mathcal{X}(S_k)$ encompassing all feasible decisions in pre-decision state $S_k$. There are two types of decisions:
\begin{enumerate}
    \item \textit{Probability display.} 
    In case of a request arrival ($r_k\neq0$), the provider displays a vector of probabilities $P_k=(p_j^k)_{j\in\mathcal{J}^P}$ with $p_{j}^k\in[0,1]$ for the probabilistic service options.
    \item \textit{Allocation.} 
    If $t_k\in\mathcal{T}^A$, the provider allocates service outcomes to the relevant pending orders $\mathcal{O}_k^A=\{o\in\mathcal{O}_k: t^A_o=t_k\}$. For orders with $j_o\in\mathcal{J}^P$, the binary decision variable $a_{o}^k$ equals 1 if order $o\in\mathcal{O}^A_k$ receives the premium outcome $i_{j_o}^+$ of the chosen probabilistic service option $j_o$, and 0 otherwise. Customers who chose a deterministic option $j_o\in\mathcal{J}^D$ directly receive the associated outcome $i_j$, and we set $a_o^k=1$. Due to the structure of service options, allocation decisions $A_k=(a_o^k)_{o\in\mathcal{O}^A_k}$ can be decomposed by premium outcome $i\in\mathcal{I}^+$ and are constrained by the respective available capacity $q_i^k$.
\end{enumerate}

Although the two decision spaces are strongly interrelated (Section~\ref{solution_approach}), we assume that they are decoupled in the sense that displayed probabilities only affect the option chosen by the customer. Accordingly, it suffices to track the customer's choice in the state variable, without storing the display that led to that choice. In particular, the provider does not allocate the premium outcome in case of insufficient capacity even when the originally displayed probability equals one, nor does the allocation decision space compel the provider to allocate the base outcome when the displayed probability is zero. Instead, we ensure consistency between decision types through transparency constraints (Section~\ref{sdp_transparency}). To avoid concurrent probability display and allocation in a single decision epoch, we assume $r_k=0$ if $t_k\in\mathcal{T}^A$ without loss of generality. 

\subsubsection{Transitions and Exogenous Information} \label{sdp_transitions} 
The exogenous information $W_{k+1}=(j_{r_k},\mathcal{C}_{k+1},r_{k+1},t_{k+1})\in\mathcal{W}$ is revealed during the transition from decision epoch $k$ to $k+1$, i.e., after the decision $X_k$. It follows a known stochastic process with probabilities $\mathbb{P}(W_{k+1}|S_k,X_k)$ and represents the three sources of uncertainty in our problem:
\begin{enumerate}
    \item \textit{Customer choice.} The customer belonging to request $r_k\neq0$ associates utilities $u_i^k$ with the service outcomes $i\in\mathcal{I}$, which are private information not observed by the provider. Based on the provider's announced probabilities $P_k$, the customer forms valuations $U^k_j=U(P_k,u_{i_j^-}^k,u_{i_j^+}^k)$ for the probabilistic service options $j\in\mathcal{J}^P$, with function $U$ reflecting the customer's risk preference. The valuation for deterministic options $j\in\mathcal{J}^D$ equals $U^k_j=u_{i_j}^k$. The customer chooses the option with the highest valuation $j_{r_k}=\argmax_{j\in\mathcal{J}}U_j^k$. After revealing the customer's choice $j_{r_k}$, the request turns into an order, yielding an updated set of pending orders $\mathcal{O}_{k+1}$. 
    In case of no request ($r_k=0$), we set $j_{r_k}=0$.
    \item \textit{Capacity utilization.} Since usage durations are stochastic (and may even take on the value zero to model no-shows), the capacity utilization $\mathcal{C}_{k+1}$ in decision epoch $k+1$ is not merely the result of a preceding allocation decision $X_k$ (if $t_k\in\mathcal{T}^A$), but is also affected by exogenous information about capacity units becoming available again.
    \item \textit{Request arrival (or lack thereof).} The subsequent decision epoch $k+1$ is either triggered by a new request arrival ($r_{k+1}\neq0$) or an allocation decision ($r_{k+1}=0$), determining the next point in time $t_{k+1}$.
\end{enumerate}
State transitions after allocation decisions additionally include a deterministic update of the pending orders $\mathcal{O}_{k+1}$ by removing allocated orders $\mathcal{O}_k^A$ from $\mathcal{O}_k$.

\subsubsection{Reward} \label{sdp_reward}
We express rewards as a function $R(S_k, X_k, W_{k+1})$. Depending on the objective, rewards may arise from the customer's choice revealed in $W_{k+1}$, e.g., if the provider aims to maximize the revenue earned from the sale of service options. Alternatively, rewards may result from allocation decisions if the specific allocated service outcome is relevant, e.g., to maximize customer satisfaction. In our computational study in Section~\ref{comp_study}, both the displayed probabilities and allocation decisions shape the reward signal based on customers' expected conditional utilities.

\subsubsection{Policy and Objective Function} \label{sdp_objective}
The solution to a sequential decision process is a policy $\pi\in\Pi$ that maps states to decisions, denoted by $X^\pi(S_k)$. The optimal policy $\pi^*$ maximizes the objective, i.e., the expected average reward per decision epoch $\bar{R}^\pi$ \citep[][Chapter 10.3]{sutton&barto_2018}:
\begin{equation} \label{eq_objective}
    \max_{\pi\in\Pi}\bar{R}^\pi = \max_{\pi\in\Pi}\left\{\lim_{K\rightarrow\infty}\frac{1}{K}\sum_{k=1}^K\mathbb{E}^\pi\left[R(S_k,X^\pi(S_k),W_{k+1})|S_0\right]\right\}
\end{equation}

\subsection{Post-Choice State and Truthful Probabilities} \label{SDP_ground_truth}
To compute truthful allocation probabilities, we introduce the concept of a \textit{post-choice state} $S_k^c\in\mathcal{S}^c$ that artificially splits the transition between two consecutive pre-decision states into two parts: We first observe the customer's choice $W_{k+1}^c=j_{r_k}$ from the outcome space $\mathcal{W}^c$, leading to the post-choice state $S_k^c=S^{MW^c}(S_k,X_k,W_{k+1}^c)$ with transition function $S^{MW^c}$. Subsequently, the remaining exogenous information $W_{k+1}^s=(\mathcal{C}_{k+1},r_{k+1},t_{k+1})\in\mathcal{W}^s$ shapes the transition to the next pre-decision state $S_{k+1}=S^{MW^s}(S_k^c,W^s_{k+1})$ with transition function $S^{MW^s}$. Compared to $S_k=(t_k,r_k,\mathcal{O}_k,\mathcal{C}_k)$, the post-choice state $S_k^c=(t_k,r_k,\mathcal{O}_{k+1},\mathcal{C}_k,\mathcal{O}_k^A,A_k)$ already contains the updated set of pending orders $\mathcal{O}_{k+1}$ (Section~\ref{sdp_transitions}) and additional information about fulfilled orders $\mathcal{O}_k^A$ together with the allocated service outcomes encoded by $A_k$ (Section~\ref{sdp_decision}) in case of a prior allocation decision ($\mathcal{O}_k^A =A_k= \{\}$ if $t_k\notin\mathcal{T}^A)$. Essentially, this concept allows us to capture the first and last moment an order exists as a pending order in the system. 

Based on this definition, the `truthful' allocation probabilities $\psi_{o}^\pi(S_k^c)$ represent the likelihood that the premium outcome $i_{j_o}^+$ will get allocated to order $o$ with $j_o\in\mathcal{J}^P$ under policy $\pi$ when the system currently occupies post-choice state $S_k^c$. Keeping in mind that $S_{k+1}=S^{MW^s}(S_k^c,W^s_{k+1})$ and $S_{k+1}^c=S^{MW^c}(S_{k+1},X^\pi(S_{k+1}),W_{k+2}^c)$, they are formalized with a recursive equation: 
\begin{align} 
    \psi_{o}^\pi(S_k^c) = 
    \begin{cases}
        \displaystyle\sum_{W_{k+1}^s\in\mathcal{W}^s}\mathbb{P}(W_{k+1}^s|S_k^c)\sum_{W_{k+2}^c\in\mathcal{W}^c}\mathbb{P}(W_{k+2}^c|S_{k+1},X^\pi(S_{k+1}))\psi^\pi_{o}(S_{k+1}^c) & \text{if } o\in\mathcal{O}_{k+1} \\[20pt]
        a_o^k & \text{if } o\in\mathcal{O}_k^A
    \end{cases} \label{eq_truthful_probabilities}
\end{align}

From each individual order's perspective, we obtain a finite-horizon stochastic process that is embedded within the overarching infinite-horizon sequential decision process. Its initial state equals the post-choice state at which the order first arises as a result of the customer choosing a probabilistic service option, and the terminal states correspond to the order's potential service outcomes.

The value function $V(S)$ represents the expected sum of rewards (adjusted by $\bar{R}^{\pi^*}$) if we start in pre-decision state $S\in\mathcal{S}$ and apply the optimal policy $\pi^*$ from that point onward \citep{mahadevan_1996}. We denote its counterpart for the post-choice states $S\in\mathcal{S}^c$ by $V^c(S)$. To accommodate the artificially split transition to and from the post-choice state, we let $R^c(S_k,X_k,W_{k+1}^c)$ and $R^s(S_k^c,W_{k+1}^s)$ denote the reward functions associated with the two respective transition stages, yielding an adapted Bellman equation:
\begin{align} 
    V(S_k) & = \max_{X_{k}\in\mathcal{X}(S_{k})}\left\{\sum_{W_{k+1}^c\in\mathcal{W}^c}\mathbb{P}(W_{k+1}^c|S_{k},X_k) \bigg( R^c(S_{k},X_{k},W_{k+1}^c) + V^c(S_{k}^c)\bigg)\right\} - \bar{R}^{\pi^*} \label{value_predecision} \\
    V^c(S_k^c) & =\sum_{W_{k+1}^s\in\mathcal{W}^s}\mathbb{P}(W_{k+1}^s|S_k^c)\left( R^s(S_{k}^c,W_{k+1}^s)+V(S_{k+1})\right) \label{value_postchoice}
\end{align}
The notation allows rewards to arise during either stage of the decomposed transition. If the original reward function $R(S_k,X_k,W_{k+1})$ jointly depends on $W_{k+1}^c$ and $W_{k+1}^s$, we can set $R^c(S_k,X_k,W_{k+1}^c)=0$, include $W_{k+1}^c$ in the post-choice state $S_k^c$, and assign the full reward to the second stage $R^s(S_{k}^c,W_{k+1}^s)$. 

\subsection{Transparency Constraints} \label{sdp_transparency}
To limit the extent to which the provider can manipulate displayed probabilities, we introduce transparency constraints. These constraints apply to all pre-decision states in which a request arrives, which we capture with a subset of the pre-decision state space $\mathcal{S}^p=\{S_k\in\mathcal{S}:r_k\neq0\}$. We let $P^\pi(S_k)=(p_{j}^\pi(S_k))_{j\in\mathcal{J}^P}$ denote the probabilities displayed under policy $\pi$ for pre-decision state $S_k$. Because the truthful probabilities defined in Section~\ref{SDP_ground_truth} refer to post-choice states, we introduce $o_{kj}$ as the hypothetical order that would arise if the request in $S_k$ chooses probabilistic service option $W_{k+1}^c=j$, yielding the hypothetical post-choice state $S_{kj}^c=S^{MW^c}(S_k,P^\pi(S_k),j)$ and the corresponding truthful probability $\psi_{o_{kj}}^\pi(S_{kj}^c)$. 

The level of intransparency $\tau\in[0,1]$ determines the maximum allowed deviation between the displayed probability $p_{j}^\pi(S_k)$ and the truthful probability $\psi_{o_{kj}}^\pi(S_{kj}^c)$ for the policy $\pi$. Accordingly, the transparency constraints for the sequential decision process encapsulated by (\ref{eq_objective}) are formalized as follows:
\begin{align} 
    &\left|p_{j}^\pi(S_k)-\psi_{o_{kj}}^\pi\left(S_{kj}^c\right)\right|\leq\tau && j\in\mathcal{J}^P,S_k\in\mathcal{S}^p\label{transparency_constraint1}
\end{align}

While $\tau=0$ represents the case of full transparency without any deviation from the truth, $\tau=1$ effectively deactivates the constraint. Unlike standard constraints that characterize the decision space (Section~\ref{sdp_decision}), the feasibility of a decision under the transparency constraints not only depends on the current decision epoch, but also on the expected downstream evolution of the sequential decision process as well as anticipated future decisions induced by the policy, which is all captured by $\psi_{o_{kj}}^\pi(S_{kj}^c)$.

The level of intransparency $\tau$ affects the optimal objective value of the constrained sequential decision process defined by (\ref{eq_objective}) and (\ref{transparency_constraint1}) as follows:\\[-0.25cm]

\textbf{Property 1:} \textit{Monotonicity in the level of intransparency.} The sets of feasible policies are nested in $\tau$; that is, $\Pi(\tau_1)\subseteq\Pi(\tau_2)$ for any $\tau_1\leq\tau_2$. Hence, the optimal expected average reward $\bar{R}^{\pi^*}(\tau)$ is nondecreasing in the level of intransparency $\tau$, pointing to an inherent incentive to distort probabilities.

\section{Solution Approach} \label{solution_approach}
Our sequential decision process is characterized by combinatorial state spaces, a continuous probability decision space, combinatorial allocation decisions, and policy-dependent transparency constraints, hindering the exact computation of the optimal policy $\pi^*$ with conventional dynamic programming methods. For a fixed policy $\pi$ and a given post-choice state $S_k^c$, we could, in principle, compute the truthful probabilities $\psi_{o}^\pi(S_k^c)$ by recursively enumerating all state transitions until the considered order $o$ is allocated, essentially generating a finite tree. Its size, however, grows combinatorially with the problem size, making the exact evaluation computationally prohibitive for realistic instances.

Constructing a policy for the problem at hand requires two components, one for probability display and one for allocation, reflecting the two types of decisions. These components are interrelated: Probability displays influence customers’ choices and therefore the composition of future orders to be allocated. Conversely, allocation decisions shape the available capacity and control the eventually assigned service outcome. In combination, both determine the truthful allocation probabilities induced by the policy. Joint optimization therefore requires the simultaneous specification of probability display and allocation decisions while accounting for the truthful probabilities endogenously generated by the overall policy to ensure feasibility with respect to the transparency constraints. 

In response, we decompose the problem by decision type. For the allocation component, we restrict our attention to \textit{non-anticipatory} allocation schemes. We characterize an allocation scheme as non-anticipatory when each allocation decision $A_k$ in states $S_k\in\mathcal{S}:t_k\in\mathcal{T}^A$ is independent of pending orders $o'\in\mathcal{O}_k:t_{o'}^A>t_k$ to be allocated at a later point in time and establish the following property (proof in Appendix~\ref{proofs}):\\[-0.25cm]

\textbf{Property 2:} \textit{Uniqueness of fully transparent probability displays under non-anticipatory allocation.} Consider a fixed, fully specified, and non-anticipatory allocation scheme $\alpha$ that uniquely determines the joint allocation outcome for all orders $o\in\mathcal{O}_k^A$ in every pre-decision state $S_k\in\mathcal{S}:t_k\in\mathcal{T}^A$ independent of pending orders $o'\in\mathcal{O}_k:t_{o'}^A>t_k$. Under full transparency ($\tau=0$), $\alpha$ admits exactly one feasible and hence optimal solution for the displayed probabilities, invariant to the underlying reward function.\\[-0.25cm]

Two consequences follow from Property~2: First, the joint policy optimization problem for $\tau=0$ within the class of non-anticipatory allocation schemes reduces to selecting an allocation scheme $\alpha$. Second, a non-anticipatory allocation scheme implies the existence of a feasible policy for $\tau=0$ and by the nesting of $\Pi(0)\subseteq\Pi(\tau)$ (Property~1), the feasible set $\Pi(\tau)$ is nonempty for every $\tau=[0,1]$. 

In particular, we employ priority rules based on order characteristics (Section~\ref{comp_study}) and hold the resulting non-anticipatory allocation schemes fixed across transparency levels $\tau$. This enables us to isolate the effect of manipulating probabilistic information as a demand-steering mechanism because any observed differences in metrics across transparency levels can be fully attributed to the probabilities displayed.

As a solution strategy, we develop an approximation-based approach. Specifically, instead of computing $\psi_{o}^\pi(S_{k}^c)$ for each individual order $o$ and post-choice state $S_{k}^c$, we train a neural network to predict approximated truthful probabilities $\hat{\psi}_{o}(S_k^c)$ and propose a corresponding learning algorithm (Section~\ref{s_approx_truth}). To efficiently search the continuous probability display decision space, we formulate a scenario-based mixed-integer program (MIP, Section~\ref{s_aopt}).

\subsection{Approximation of Truthful Probabilities} \label{s_approx_truth}
We first outline the architecture of the neural network in Section~\ref{network_architecture}. Following a detailed description of our learning algorithm in Section~\ref{learning_algorithm}, we highlight important algorithmic design choices in Section~\ref{design_of_updates}. Lastly, we explain how to evaluate the approximation quality in Section~\ref{accuracy_evaluation}.

\subsubsection{Approximation Architecture} \label{network_architecture}
Our neural network $\hat{\psi}_{\bm{\theta}}$ is parameterized by weights ${\bm{\theta}}$ that are learned offline in a simulation. The network takes a vector of features ${\bm{\phi}}(S_k^c,o)$ as an input that encodes information about the post-choice state $S_k^c$ and the (hypothetical or pending) order $o$ for which we aim to predict the truthful probability $\hat{\psi}_{o}(S_k^c)$. We normalize all features to the interval $[-1,1]$ to mitigate differences in scale and prevent features with larger numerical magnitudes from dominating the network's weight updates. The output layer consists of a single neuron with an activation function whose range is bounded by $[0,1]$, ensuring that the network output can be interpreted as a probability. For a newly arrived request, we conduct one forward pass $\hat{\psi}_{\bm{\theta}}({\bm{\phi}}(S_{kj}^c,o_{kj}))$ through the neural network per probabilistic service option $j\in\mathcal{J}^P$, each with the respective hypothetical order.

We explain the concrete features and detailed network architecture used for our computational study in Appendix~\ref{policy_details}. While the design of the features ${\bm{\phi}}(S_{kj}^c,o_{kj})$ is inherently application-specific, we provide general conceptual guidance: Natural choices include the amount of time remaining until the order's allocation, the number of pending orders for probabilistic service option $j$, the presently occupied capacity of the associated premium outcome $i_j^+$ along with elapsed usage durations, and attributes of the request and the probabilistic service option under consideration. We also include the expected demand share of the premium outcome $i_j^+$ as a measure of its popularity. 

Beyond this readily available information, we construct more sophisticated features with insights gained from a lookahead approach, inspired by the corresponding policy class in the framework by \cite{powell_2022}. Specifically, we simulate several sample paths starting from $S_{kj}^c$ under a fixed, state-independent base policy to obtain rough estimates of the capacity expected to be available upon allocation as well as the truthful probability. Because the base policy remains fixed over time and its decisions do not depend on specific pre-decision states, most computations for the lookahead procedure can be performed in advance and reused efficiently. The neural network then learns to correct the biased baseline estimate towards the actual policy. 

\subsubsection{Learning Algorithm} \label{learning_algorithm}
At first glance, approximating the truthful probabilities may resemble a standard supervised learning task. However, these probabilities depend on the policy, while the policy itself is in turn linked to the probabilities through the transparency constraints, resulting in a feedback loop. To address this endogeneity, we develop an iterative, simulation-based learning procedure inspired by reinforcement learning \citep{sutton&barto_2018} that is conducted offline. Like value function approximation, our method replaces a quantity that depends on the downstream evolution of the sequential decision process, namely $\psi^\pi_{o}(S_{k+1}^c)$ in (\ref{eq_truthful_probabilities}), with a function that generalizes across states. The key difference is that we approximate the conditional distribution of a future outcome given the current post-choice state and the policy, yielding an approximation strategy that targets the sequential decision problem's outcome space instead of the value or decision space.

The training procedure is summarized in Algorithm \ref{alg_training}. Whenever a request arrives in a simulated decision epoch $n$, we make a probability display decision. More precisely, a weighted coin flip determines whether we exploit our current knowledge encapsulated by $\hat{\psi}_{\bm{\theta}}$ and solve the scenario-based MIP described in Section~\ref{s_aopt} or explore by randomly sampling a decision from the decision space. The probability of exploring is high at the beginning and exponentially decreases over the course of training. This exploration mitigates a self-reinforcing sampling bias that may arise from the interaction between the probability approximation and customers' choices: If the network underestimates the truthful probability for a probabilistic service option, the resulting policy displays a correspondingly low probability. Customers may then select this option only rarely, generating few orders and, consequently, few allocation outcomes from which the underestimation could be corrected. Exploratory display decisions increase the likelihood of obtaining observations for otherwise underrepresented probabilistic options and outcomes.

During the simulation, we collect experiences in a replay memory $\mathcal{E}$, which stores the training data for updating the network weights ${\bm{\theta}}$. After a probability display decision, we generate new experiences $e$ in the form of tuples $(o_e,{\bm{\phi}}_e,a_e)=(o,{\bm{\phi}}(S_{n}^c,o),\cdot)$ from each pending order $o\in\mathcal{O}_n$ based on the order's feature vector in the post-choice state of the current decision epoch ${\bm{\phi}}_e={\bm{\phi}}(S_{n}^c,o)$. Once an order $o$ is allocated, we update all experiences $e\in\mathcal{E}:o_e=o$ associated with that order by including the realized allocation outcome $a_e=a_o^n$. We deliberately create experiences for all pending orders, rather than only for the order newly confirmed in the current decision epoch, because hypothetical and actual pending orders provide equivalent training observations for the network.

Given that the observed allocation outcomes are binary and we want to predict the underlying probability, our learning problem closely resembles a binary classification task. Accordingly, we train the network by minimizing the binary cross-entropy loss. To this end, we sample a mini-batch $\mathcal{B}$ of experiences from the replay memory for which the allocated service outcome $a_e$ has already been observed and update the network parameters by performing a gradient-based optimization step on the loss function:
\begin{align}
\mathcal{L}({\bm{\theta}},\mathcal{B})=-\frac{1}{|\mathcal{B}|}\sum_{e\in\mathcal{B}}\left(a_e\log\hat{\psi}_{{\bm{\theta}}}({\bm{\phi}}_e)+(1-a_e)\log(1-\hat{\psi}_{{\bm{\theta}}}({\bm{\phi}}_e))\right) \label{loss_function}   
\end{align}

\begin{algorithm}
\fontsize{10}{14}\selectfont
\caption{Offline training of network weights ${\bm{\theta}}$}\label{alg_training}
    \textbf{Initialization:} Network $\hat{\psi}_\mathbf{\theta}$ with random weights ${\bm{\theta}}$, replay memory $\mathcal{E}=\{\}$, initial pre-decision state $S_1$, $e=0$
\begin{algorithmic}[1]
\For{$n=1,\dots,N$}
    \If{$r_n\neq 0$}
        \If{exploration}
             \State uniformly sample $P_n$ from $\mathcal{X}(S_n)$
        \Else
            \For{probabilistic product $j\in\mathcal{J}^P$}
                \State compute features ${\bm{\phi}}(S_{nj}^c,o_{nj})$ for hypothetical order $o_{nj}$
                \State approximate truthful probabilities $\hat{\psi}_j^n=\hat{\psi}_{\bm{\theta}}({\bm{\phi}}(S_{nj}^c,o_{nj}))$
            \EndFor
            \State make probability display decision $P_n$ with MIP (Section \ref{s_aopt}) based on $\hat{\psi}_j^n$ for $j\in\mathcal{J}^P$
        \EndIf
        \For{pending order $o\in\mathcal{O}_n$}
            \State $e\leftarrow e+1$
            \State generate experience $(o_e,{\bm{\phi}}_e,a_e)=(o,{\bm{\phi}}(S_{n}^c,o),\cdot)$ and store it in replay memory $\mathcal{E}$
        \EndFor
    \Else
        \State make allocation decision
        \For{allocated order $o\in\mathcal{O}_n^A$}
            \State update relevant experiences $e\in\mathcal{E}:o_e=o$ by setting $a_e$ to allocation outcome $a_o^n$
        \EndFor
    \EndIf
    \State update network weights ${\bm{\theta}}$
    \State sample exogenous information $W_{n+1}$ and transition to $S_{n+1}$
\EndFor
\end{algorithmic}
\end{algorithm}

\subsubsection{Design Choices} \label{design_of_updates}
We highlight two distinctive design choices for our learning algorithm that proved to be crucial for performance:
\begin{enumerate}
    \item \textit{No temporal difference learning.} From a reinforcement learning perspective, the embedded finite stochastic process associated with an order constitutes an approximation task characterized by highly sparse, delayed, and binary `reward' signals: For each order, the allocated service outcome is observed only once as a binary signal at the time of allocation, with no intermediate feedback from the simulation environment. As a result, we found that temporal difference learning methods, which exploit the recursive structure in (\ref{eq_truthful_probabilities}) and update the network parameters using the predicted truthful probability in the subsequent state, lead to unsatisfactory performance. Instead, we employ a Monte Carlo-style update \citep[][Chapter 5]{sutton&barto_2018} built on the observed allocation decision to compute the loss (\ref{loss_function}). Although basing the update on complete trajectories increases the variance, it ensures that each update is grounded in an observed outcome.
    \item \textit{Sampling of experiences.} Instead of drawing the mini-batch $\mathcal{B}$ uniformly and independently, we first sample a set of previously encountered post-choice states represented in the replay memory. For each sampled state, we include all associated orders with observed allocation outcomes in the update, grouping them into several batches if their number exceeds $|\mathcal{B}|$. Incorporating pending orders associated with the same state jointly in the update is essential, as it allows the network to learn to differentiate between orders in the same state that ultimately experience different allocation decisions.
\end{enumerate}

\subsubsection{Evaluating Approximation Quality} \label{accuracy_evaluation} 
By design, any reinforcement learning algorithm is unlikely to achieve predictive accuracy across the entire state space and for all hypothetical orders in our setting. In particular, an identification problem arises for combinations of requests and service options that are never observed in the offline simulation. For example, if some customers consistently prefer a deterministic option to certain probabilistic service options, the simulation never generates allocation decisions for this combination of customer and service option, leading to a lack of training data. 

At the same time, prediction errors for such combinations have limited managerial relevance because customers likewise cannot evaluate the accuracy of probabilities displayed for service options they do not choose. Instead, long-term credibility depends both on the approximation quality for chosen options and the extent to which the provider manipulates disclosed information. Accordingly, the central question is whether the displayed probabilities appear trustworthy to customers over repeated interactions, given the allocation outcomes they actually observe, instead of counterfactual accuracy for service options not selected.

To assess the quality of the learned approximation, we generate a test dataset and restrict attention to the chosen service options. Because the exact truthful probabilities are unavailable for comparison, we estimate them via Monte Carlo simulation for post-choice states randomly selected during an independent simulation. Specifically, for each sampled post-choice state $S_k^c$, we simulate $F$ trajectories under the given policy, each starting from $S_k^c$ and continuing until the allocation decisions for all relevant pending orders with $j_o\in\mathcal{J}^P$ have been observed. For each trajectory $f=1,\ldots,F$, we encode by $a_{of}\in\{0,1\}$ whether pending order $o$ received the premium outcome $i_{j_o}^+$ of the chosen probabilistic service option $j_o\in\mathcal{J}^P$. The Monte Carlo estimate of the truthful probability is then given by $\hat{\psi}^{MC}_{o}(S_k^c)=\frac{1}{F}\sum_{f=1}^Fa_{of}$. We compare these estimates with the corresponding neural network predictions $\hat{\psi}_{o}(S_k^c)$ and assess approximation quality using the mean absolute error across the evaluated state-order combinations. Note that because the Monte Carlo estimates are based on a finite number of trajectories, they are themselves subject to sampling error.

Importantly, this procedure allows us to evaluate probability predictions for pending orders at arbitrary stages, rather than only immediately after order confirmation. If desired, the evaluation can be restricted to newly confirmed orders by filtering the test data accordingly.

\subsection{Probability Display} \label{s_aopt}
To decide which probabilities to display in a pre-decision state $S_k$, the provider must essentially evaluate the maximization operator in (\ref{value_predecision}) subject to the transparency constraints (\ref{transparency_constraint1}). Similar to the truthful probabilities, the exact value $V^c(S_{k}^c)$ of a post-choice state is generally intractable to compute for realistic instances, so we likewise replace it with an approximation $\hat{V}^c(S_{k}^c)$. To avoid nonlinearity, the approximations should only depend on the option selected by the customer and not on the specific probabilities $P_k$ displayed prior to that choice. Reusing the notation introduced in Section~\ref{sdp_transitions}, we write $\hat{V}^c(S_{kj}^c)$ for the approximate value of the hypothetical post-choice state resulting from the current request choosing service option $j\in\mathcal{J}$.

Our methodology is compatible with any of the commonly applied value approximation strategies, spanning both online methods, such as a truncated rollout or lookahead model, and offline, learning-based methods; we refer to \cite{powell_2022} and \cite{sutton&barto_2018} for comprehensive introductions. Accordingly, the specific approach to obtain $\hat{V}^c(S_{kj}^c)$ is not the focus of this work. We describe the particular methods used for our computational study in Section~\ref{comp_study}.

In addition to the value approximation $\hat{V}^c(S_{kj}^c)$, solving the maximization problem requires a tractable closed-form representation of the choice probabilities $\mathbb{P}(W_{k+1}^c|S_{k},X_k)$ that captures their dependence on the continuous probability decision variables $P_k$ (encapsulated by $X_k$). Because such a representation is not generally available, we develop a scenario-based approach inspired by \cite{paneque_etal_2021}. Specifically, we generate scenarios $\omega\in\Omega$ by sampling the service outcome utilities $\Tilde{u}_{i\omega}$ for the current customer request from their known distributions. For deterministic products $j\in\mathcal{J}^D$, this immediately yields the valuation $\Tilde{U}_{j\omega}=\Tilde{u}_{i_j\omega}$ as a constant parameter since the base and premium outcome coincide. For probabilistic service options $j\in\mathcal{J}^P$, the probability display decision $P_k$ and the scenario $\omega \in \Omega$ jointly determine the valuation, resulting in a continuous auxiliary variable $\Tilde{U}_{j\omega}$. The customer selects the option with the highest valuation. We capture the maximum valuation in scenario $\omega$ with the continuous variable $\widetilde{U}_{\omega}$ and represent the customer's choice with the binary decision variable $c_{j\omega}$, which equals one if the customer selects service option $j$ in scenario $\omega$ and zero otherwise. The corresponding choice probability is approximated by the sample frequency $\hat{\mathbb{P}}( W_{k+1}^{c}=j \mid S_k,X_k ) = \frac{1}{|\Omega|} \sum_{\omega\in\Omega} c_{j\omega}$.

Piecing everything together, we obtain the following mixed integer program:
\begin{align}
    & \max \sum_{j\in\mathcal{J}}\frac{1}{|\Omega|}\sum_{\omega\in\Omega}c_{j\omega}\bigg( R(S_{k},P_{k},j) + \hat{V}^c(S^c_{kj})\bigg)&& \label{MIP_objective} \\
    & \Tilde{U}_{j\omega}=U\left(p_{j}^k,\Tilde{u}_{i_j^-\omega},\Tilde{u}_{i_j^+\omega}\right) && j\in\mathcal{J}^P,\omega\in\Omega \label{MIP_compute_valuations}\\
    & \Tilde{U}_\omega\geq\Tilde{U}_{j\omega} && j\in\mathcal{J},\omega\in\Omega \label{MIP_max_valuation1}\\
    & \Tilde{U}_\omega\leq\Tilde{U}_{j\omega} + M(1-c_{j\omega}) && j\in\mathcal{J},\omega\in\Omega \label{MIP_max_valuation2}\\
    & \sum_{j\in\mathcal{J}}c_{j\omega}=1 && \omega\in\Omega \label{MIP_choice}\\
    &p_{j}^\pi(S_k)-\hat{\psi}_j^k\leq\tau && j\in\mathcal{J}^P \label{MIP transparency1} \\
    &\hat{\psi}_j^k-p_{j}^\pi(S_k)\leq\tau && j\in\mathcal{J}^P \label{MIP transparency2} \\
    & p_{j}^k\in[0,1] && j\in\mathcal{J}^P \\
    & c_{j\omega}\in\{0,1\} && j\in\mathcal{J},\omega\in\Omega \\
    & \Tilde{U}_{j\omega},\Tilde{U}_\omega \in \mathbb{R} && j\in\mathcal{J}^P,\omega\in\Omega \label{MIP_domains}
\end{align}

Objective (\ref{MIP_objective}) approximates the maximum operator in (\ref{value_predecision}). Constraints (\ref{MIP_compute_valuations}) determine the scenario- and probability-dependent valuations of probabilistic service options. Constraints (\ref{MIP_max_valuation1}) and (\ref{MIP_max_valuation2}), together with a sufficiently large constant $M$, identify the customer's maximum valuation per scenario and ensure that the associated option is selected. Constraints (\ref{MIP_choice}) require the customer to choose one option in each scenario. Finally, (\ref{MIP transparency1}) and (\ref{MIP transparency2}) limit the deviation between displayed and approximated truthful probabilities with $\hat{\psi}_j^k=\hat{\psi}_{\bm{\theta}}({\bm{\phi}}(S_{kj}^c,o_{kj}))$. Note that if the immediate reward $R(S_{k},P_{k},j)$ directly depends on $P_k$ or function $U$ models nonlinear risk preferences, linearization techniques may be necessary.

There are two sources of multiple optimal solutions in our model, which we address as follows:
\begin{itemize}
    \item \textit{Indifference.} If multiple service options achieve the maximum valuation in a scenario, the customer is indifferent among them. The optimization automatically resolves such ties in favor of the provider, which not necessarily aligns with actual customer behavior. A relevant structural tie arises if the base outcome of a probabilistic service option $j\in\mathcal{J}^P$ is also offered as a deterministic option $j'\in\mathcal{J}^D$ with $i_j^-=i_{j'}$. If $p_j^k=0$, the product valuations are identical since both products are expected to yield the same outcome from the customer's perspective. We denote the subset of probabilistic service options with such a deterministic counterpart by $\mathcal{J}^{P'}$. Assuming that the customer only selects a probabilistic service option $j\in\mathcal{J^{P'}}$ if $p_j^k\geq\frac{1}{M'}$, we can formulate an additional tie-breaking constraint with a sufficiently large constant $M'$:
    \begin{align}
    & c_{j\omega}\leq M'\cdot p_j^k && j\in\mathcal{J}^{P'},\omega\in\Omega
    \end{align}
    \item \textit{Intransparency.} For $\tau>0$, different probability displays may consistently provoke the same customer choices in all scenarios and hence yield an identical objective value. To ensure that the provider distorts probabilities only as much as necessary, we penalize deviations from the approximated probabilities with a small positive constant in the objective.
\end{itemize}

\section{Computational Experiments} \label{comp_study}
Our computational study investigates how, in the context of out-of-home delivery (OOHD), displayed probabilities can serve as an information-based demand management tool and how (in)transparency affects the involved stakeholders, namely the provider and customers. 

\subsection{Design of Experiments} \label{cs_design}
In this section, we apply our general modeling framework to OOHD (Section~\ref{cs_OOHD}) and specify the instance settings (Section~\ref{cs_settings}), investigated policies (Section~\ref{cs_policies}), and evaluation metrics (Section~\ref{cs_kpi}). 

\subsubsection{The Case of Out-of-Home Delivery} \label{cs_OOHD}
OOHD providers commonly operate parcel lockers and parcel shops simultaneously within their networks since both facility types offer distinct advantages and disadvantages \citep{janinhoff_etal_2024}: While the opening hours of shops are restricted, fully automated lockers enable 24/7 access for customers. In contrast, capacity constraints at shops are negligible in practice, whereas locker capacity is strictly limited by the fixed number of compartments. Additionally, the availability of locker capacity is uncertain because it depends on customers' pickup behavior, rendering lockers a natural candidate for probabilistic service options. More precisely, we model parcel shops as base outcomes $i\in\mathcal{I}^-$ and lockers as premium outcomes $i\in\mathcal{I}^+$. Each probabilistic service option $j\in\mathcal{J}^P$ with $|\mathcal{J}^P|=|\mathcal{I}^+|$ pairs a locker with its closest parcel shop, to which the parcel gets redirected if the locker's capacity is insufficient. Customers may also directly choose a parcel shop, included as deterministic service options $j\in\mathcal{J}^D$ with $|\mathcal{J}^D|=|\mathcal{I}^-|$.

We divide the planning horizon into days and map each point in time $t_k$ to a tuple $(t_k^d,t_k^w)$ where $t_k^d=1,\dots,\infty$ denotes the current day and $t_k^w=1,\dots,T+1$ the point in time within a day. Requests may arrive at $t_k^w=1,\dots,T$ and are characterized by the customers' locations. At the end of each day $(t_k^d,t_k^w=T+1)$, the provider makes allocation decisions for all pending orders that arrived throughout day $t_k^d$.

Following random utility theory \citep{train_2009}, we decompose the utility $u_i^k$ that a request $r_k$ associates with facility $i\in\mathcal{I}$ into a deterministic and a stochastic component. The deterministic component $v_i^k=\beta_L\cdot\mathbf{1}(i\in\mathcal{I}^+) -\beta_D\delta_{r_ki}$ includes an additive constant $\beta_L>0$ for parcel lockers, reflecting their benefit of unrestricted opening hours, and linearly depends on the Euclidean distance $\delta_{r_ki}$ between the customer's location and facility $i$ with coefficient $\beta_D>0$. The stochastic component $\varepsilon_{i}^k$ follows a known distribution and captures individual preferences as well as other unobserved factors. In total, we obtain $u_i^k = v_i^k +\varepsilon_{i}^k$.

The provider's objective is to maximize aggregate customer satisfaction, measured by the sum of customers' utilities for the allocated facilities. Although the provider does not observe utilities $u_i^k$ directly, the displayed probability vector $P_k$ and the chosen service option $j_{r_k}$ provide information about them. Once request $r_k$ becomes an order $o$, we store the conditional expected utilities $\mathbb{E}[u^k_i|P_k,j_{r_k}]$ for service outcomes $i\in\{i_{j_{r_k}}^-,i_{j_{r_k}}^+\}$ as order attributes $\hat{u}_o^-,\hat{u}_o^+$ in the state variable. For allocation decisions, the reward function thus equals $R(S_k,A_k,W_{k+1})=\sum_{o\in\mathcal{O}_k^A}(a_o^k\hat{u}_o^++(1-a_o^k)\hat{u}_o^-)$ with $R(S_k,P_k,W_{k+1})=0$ in case of probability display.

\subsubsection{Settings} \label{cs_settings}
We begin by specifying the parameters held constant across all settings:
\begin{itemize}
    \item \textit{Infrastructure.} Following \cite{akkerman_etal_2025}, we use the RC benchmark instance with 600 customer locations by \cite{gehring&homberger_2002} and randomly designate a setting-dependent number of locations as OOHD facilities. Of these facilities, $80\%$ are parcel lockers with $Q_i=10$ compartments each, and $20\%$ are parcel shops. The remaining locations represent the customer pool. We rescale the coordinates to an area of $10\times 10$ km.
    \item \textit{Requests.} At each within-day point in time $t_k^w\in\{1,\ldots,T\}$, a request arrives with probability $0.9$, and its location is sampled uniformly from the customer pool. 
    \item \textit{Utilities.} We set the distance-related utility coefficient to $\beta_D=4$. In line with the multinomial logit model \citep{train_2009}, the stochastic utility components $\varepsilon_{i}^k$ follow independent standard Gumbel distributions. We assume risk-neutral customers such that the valuations $U_j^k$ for probabilistic service options $j\in\mathcal{J}^P$ equal the expected utility $U_j^k=p_j^ku_{i_j^+}^k+(1-p_j^k)u_{i_j^-}^k$. Because the conditional expectations $\mathbb{E}[u^k_i|P^k,j_{r_k}]$ cannot be computed in closed form, we estimate them through accept-reject sampling \citep{train_2009} using 100 accepted draws.
    \item \textit{Pickup behavior.} Customers collect their parcels on the first, second, or third day after allocation with probabilities of 60\%, 20\%, and 20\%, respectively. The pickup time within each day is distributed uniformly.
\end{itemize}

Starting from a \textit{base setting} with the additive locker utility constant set to $\beta_L=2$ and $I=10$ facilities, we vary the experimental setup along two dimensions:
\begin{enumerate}
    \item We examine different levels of systematic utility difference between lockers and shops $\beta_L\in\{0,4\}$.
    \item To assess the scalability of our approach, we increase the number of OOHD facilities to $I\in\{15,20,25\}$ while maintaining a locker share of $80\%$. The resulting facility networks are nested, i.e., each larger network contains all facilities of the smaller networks. Throughout our study, we scale demand proportionally with network size by setting the number of within-day periods to $T=6I$.
\end{enumerate}

\subsubsection{Policies} \label{cs_policies}
We test two non-anticipatory allocation schemes. In both cases, we rank the orders $o\in\mathcal{O}_k^A:i_{j_o}^+=i$ for each parcel locker $i\in\mathcal{I}^+$ according to a specific priority criterion and allocate them to the premium outcome (parcel locker) as long as sufficient capacity is available. Remaining orders receive the base outcome (parcel shop). The priority criterion for the respective allocation scheme must be stored as an additional order attribute in the state variables.
\begin{enumerate}
    \item \textbf{FF} (\textit{first come, first served}): As a benchmark, we process the orders according to their arrival time.
    \item \textbf{MY} (\textit{myopic}): Following a regret-based rationale, we rank the orders in descending order of the difference between the deterministic utility components $v_i^k$ for locker and shop. Essentially, this prioritizes customers who incur the largest deterministic utility loss if redirected to the parcel shop. We deliberately refrain form ranking orders based on their conditional expected utilities $\hat{u}_o^+$ and $\hat{u}_o^-$ because these quantities depend on the specific probabilities displayed upon request arrival, which would create a feedback loop between probability display and allocation decisions.
\end{enumerate}

For probability display, we consider two variants of our approach that differ in the technique to obtain the approximated post-choice state values $\hat{V}^c(S_k^c)$ and compare them to two benchmarks:
\begin{enumerate}
    \item \textbf{EU} (\textit{expected utility}): Focusing entirely on the current request, we approximate $\hat{V}^c(S_k^c)$ by the hypothetical order's expected future reward. Based on the estimated truthful probabilities $\hat{\psi}_j^k$ and the sampled utilities $\Tilde{u}_{i \omega}$, we compute the scenario-dependent approximation $\hat{V}^c(S_{kj}^c,\omega)=\hat{\psi}_j^k\Tilde{u}_{i_j^+\omega}+(1-\hat{\psi}_j^k)\Tilde{u}_{i_j^-\omega}$.
    \item \textbf{LA} (\textit{lookahead model}): To account for the potential displacement of future customers, we generate scenarios for the future evolution of the system, solve the resulting finite-horizon full information problems, and derive $\hat{V}^c(S_{kj}^c)$ based on the objective values (details in Appendix \ref{policy_details}).
    \item \textbf{NO} (\textit{no information}): To gauge the value of information provision, we include a benchmark where the provider does not communicate probabilities to customers, effectively offering flexible products instead (Section~\ref{intro_rw_revenue_mgmt}). We assume that customers disregard allocation uncertainty in their valuations and select the service option $j\in\mathcal{J}$ whose premium outcome yields the highest utility $u_{i_j^+}^k$.
    \item \textbf{GP} (\textit{guaranteed premium}): As an alternative to probabilistic service options, the provider may decide to offer a premium outcome only if it can be guaranteed, entirely eliminating the case of parcels being redirected to shops. This requires checking whether a feasible allocation that includes all pending orders exists assuming that customers collect their parcels as late as possible. The displayed probability equals 1 if allocating the request to the premium outcome is feasible under this worst-case assumption and 0 otherwise.
\end{enumerate}

Combining the approaches listed above yields $2\cdot 4=8$ policies. We denote each policy by $\pi$, with probability display indicated in the superscript and allocation in the subscript, e.g., $\pi^\text{EU}_\text{FF}$. For policies compatible with transparency constraints (EU and LA), we include the level of intransparency, e.g., $\pi^\text{EU}_\text{FF}(\tau)$. Across all settings, we consider both full transparency ($\tau=0$), where the provider merely displays probabilities to inform customers, and completely deactivated transparency constraints ($\tau=1$), allowing the provider to freely leverage the probabilities as a demand management tool. In the setting with $\beta_L=4$, we conduct a more detailed analysis by varying $\tau\in\{0,0.05,0.1,\dots,1\}$.

We implemented all algorithms in Python~3.13 and solved the MIPs for probability display decisions to optimality with Gurobi~13 if $\tau>0$. Under full transparency $\tau=0$, the unique feasible solution to the MIP is directly given by $p_j^k=\hat{\psi}_j^k$. For EU and LA, we train the approximation $\hat{\psi}_{\bm{\theta}}$ for every setting and value of $\tau$, including $\tau=1$. Although the transparency constraints are inactive for $\tau=1$, we still require $\hat{\psi}_{{\bm{\theta}}}$ to minimize deviations from the approximated truthful probabilities $\hat{\psi}_j^k$ as a secondary objective (Section~\ref{s_aopt}). Appendix~\ref{policy_details} provides details on the features ${\bm{\phi}}(S_k^c,o)$ and hyperparameters. The experiments were conducted on a server equipped with two Intel Xeon E7-8890 v3 processors (2.5 GHz, 18 cores) and 512~GB of RAM. The average computation time for probability display and allocation decisions remained below 0.5 seconds for all considered policies and settings, indicating suitability for real-time deployment. 

\subsubsection{Metrics} \label{cs_kpi}
We use eight metrics to evaluate solution quality and derive managerial insights:

\begin{enumerate}
    \item \textbf{CS} (\textit{customer satisfaction}): As our primary performance metric, we report customer satisfaction, measured by the conditional expected utilities for allocated service outcomes.
    \item \textbf{LS} (\textit{locker share}): The share of customers that select a probabilistic service option.
    \item \textbf{RR} (\textit{redirection rate}): The percentage of customers receiving a parcel shop base outcome out of those that chose a probabilistic service option.
    \item \textbf{LU} (\textit{locker utilization}): The percentage of occupied locker capacity directly after allocation.
    \item \textbf{DC} (\textit{displayed probability for chosen option}): The probability displayed for the premium outcome if a customer selects a probabilistic service option.
    \item \textbf{IN} (\textit{level of intransparency}): The maximum absolute deviation between the displayed probabilities $p_j^\pi(S_k)$ and the approximated truthful probabilities $\hat{\psi}_{o}(S_k^c)$ per customer.
    \item \textbf{AE} (\textit{approximation error}): The mean absolute error between our neural network's predictions $\hat{\psi}_{o}(S_k^c)$ and the Monte Carlo estimates $\hat{\psi}_o^{MC}(S_k^c)$.
    \item \textbf{CR} (\textit{customer regret}): The utility difference  between the outcome allocated under full transparency ($\tau=0$) and deactivated transparency constraints ($\tau=1$) per customer.
\end{enumerate}

Throughout our study, we report probabilities and the metrics LS, RR, LU, and DC as percentages (\%), whereas IN and AE are measured in percentage points (pp). In each setting, we estimate the metrics with 30 randomly generated instances. For the sake of comparability, we consistently use the same 30 sequences of customer arrivals (including location, stochastic utility components, and pickup behavior). The policies start with empty facilities and no pending orders and complete a warm-up phase of 25 days that are discarded to eliminate temporary effects of the initial state. We calculate the metrics per instance (consisting of 75 days after the warm-up) and compute the average per setting. To evaluate approximation quality, we randomly select 10 decision epochs from each instance and obtain the Monte Carlo estimates $\hat{\psi}^{MC}_o$ for all pending orders based on 100 sampled paths (Section~\ref{accuracy_evaluation}).

\subsection{Base Setting} \label{cs_base}
We first report aggregate results (Section \ref{cs_base_overall}), followed by a detailed analysis of the distortion of probabilities (Section \ref{cs_base_distortion}), the redistribution of demand (Section \ref{cs_base_dmnd_mgmt}), the impact of demand management on individual customers (Section \ref{cs_base_regret}), and long-term credibility (Section \ref{cs_base_credibility}).

\subsubsection{Solution Quality} \label{cs_base_overall}
Table~\ref{table_kpi_base} summarizes the metrics defined in Section~\ref{cs_kpi} (except CR, which we analyze in Section~\ref{cs_base_regret}) per policy in the base setting. For customer satisfaction (CS), we include the average aggregate value over the entire evaluation period (Total), its relative improvement over NO under the same allocation scheme in percent ($\Delta\%$), and the mean per customer (pc) to facilitate the comparison across settings. For locker utilization (LU), we report the minimum ($\underline{\text{LU}}$) and maximum ($\overline{\text{LU}}$) average utilization across all lockers.

\begin{table}
\centering
\caption{Performance metrics for the base setting.} 
{\centering
\label{table_kpi_base}
\scriptsize
\begin{tabular}{llccccccccccc}
\toprule
 & & & \multicolumn{3}{c}{CS} & \multicolumn{5}{c}{(\%)} & \multicolumn{2}{c}{(pp)} \\ \cmidrule(lr){4-6} \cmidrule(lr){7-11} \cmidrule(lr){12-13}
\multicolumn{2}{l}{Policy $\pi$} & $\tau$ & Total & $\Delta$\% & pc & LS & RR & $\underline{\text{LU}}$ & $\overline{\text{LU}}$ & DC & IN & AE \\
\midrule
FF & NO & -- & -19,996.5 & 0.0 & -4.9 & 86.6 & 13.4 & 59.8 & 96.6 & -- & -- & -- \\
 & GP & -- & -19,950.2 & 0.2 & -4.9 & 78.9 & 0.0 & 75.5 & 90.4 & -- & -- & -- \\
 & EU/LA & 0 & -17,439.2 & 12.8 & -4.3 & 85.6 & 5.5 & 69.1 & 96.7 & 94.9 & 0.0 & 0.8 \\
 & EU & 1 & -17,408.6 & 12.9 & -4.3 & 85.5 & 5.2 & 69.7 & 96.6 & 94.7 & 1.1 & 0.6 \\
 & LA & 1 & -16,551.2 & 17.2 & -4.1 & 82.3 & 2.7 & 76.3 & 92.0 & 91.4 & 19.9 & 0.4 \\ \midrule
MY & NO & -- & -18,108.0 & 0.0 & -4.5 & 86.6 & 13.4 & 59.8 & 96.4 & -- & -- & -- \\
 & GP & -- & -19,950.2 & -10.2 & -4.9 & 78.9 & 0.0 & 75.5 & 90.4 & -- & -- & -- \\
 & EU/LA & 0 & -16,741.6 & 7.5 & -4.1 & 86.5 & 8.6 & 74.4 & 95.8 & 90.0 & 0.0 & 3.4 \\
 & EU & 1 & -16,699.1 & 7.8 & -4.1 & 86.4 & 8.5 & 75.1 & 95.7 & 90.2 & 1.8 & 3.5 \\
 & LA & 1 & -15,917.5 & 12.1 & -3.9 & 84.7 & 5.5 & 72.4 & 93.3 & 89.6 & 11.1 & 2.3 \\
\bottomrule
\end{tabular}}
\end{table}

Except for GP, allocation scheme MY consistently outperforms FF in terms of customer satisfaction given that its priority criterion aligns more closely with the objective. GP produces identical results under both FF and MY because every customer who selects a locker receives the premium outcome for certain, reducing the redirection rate (RR) to zero and rendering allocation decisions trivial. On the flip side, this guarantee is very conservative, lowering the locker share (LS) from $86.6\%$ under NO to $78.9\%$. Although minimum average locker utilization ($\underline{\text{LU}}$) increases, reflecting a more balanced use of locker capacity, the maximum average utilization across lockers ($\overline{\text{LU}}$) decreases as a result of the restrictive policy. Overall, GP generates no meaningful improvement over NO under FF and even deteriorates performance under MY.

In contrast, providing truthful probabilistic information ($\tau=0$) substantially raises customer satisfaction. Since the transparency constraints for $\tau=0$ imply a unique feasible solution for the displayed probabilities, EU and LA yield identical results (Property~2). Full transparency improves CS by $12.8\%$ under FF and $7.5\%$ under MY relative to NO. Simultaneously, it lowers the RR from $13.4\%$ to $5.5\%$ and $8.6\%$, respectively, while keeping the LS nearly constant. Similar to GP, the increase in $\underline{\text{LU}}$ indicates that probability disclosure redistributes demand away from highly utilized lockers toward facilities with more available capacity.

Removing the transparency constraints ($\tau=1$) produces negligible additional gains under EU. Notably, the policy distorts probabilities only slightly by up to $1.1$ and $1.8$~pp on average (IN). Conversely, LA with $\tau=1$ achieves the highest customer satisfaction under both allocation schemes. Unlike EU, which considers only the expected utility of the current request, LA also accounts for the opportunity cost of consuming locker capacity and the potential displacement of other customers. Accordingly, it steers demand more proactively, as reflected in the lower LS and RR. At the same time, these gains are the product of probability distortion of up to $19.9$ and $11.1$~pp. Across all policies with probability display, customers tend to select options with a high displayed probability of receiving the premium outcome (DC).

Finally, approximation errors (AE) range from $0.4$ to $3.5$~pp and are consistently lower under FF than under MY. This is to be expected given that FF constitutes a simpler prediction task: Since request arrival times serve as the priority criterion, an order's ranking relative to other pending orders is fixed and cannot be altered by later-arriving orders. Uncertainty only arises from the number of customers picking up their parcels before allocation. In comparison, later-arriving orders under MY may displace earlier ones, creating an additional source of uncertainty. 

\subsubsection{Distortion of Probabilities} \label{cs_base_distortion}
\begin{table}
\centering
\caption{Probability distortion.} 
{\centering
\label{table_distortions}
\scriptsize
\begin{tabular}{lcccc}
\toprule
& \multicolumn{2}{c}{Negative} & \multicolumn{2}{c}{Positive} \\ \cmidrule(lr){2-3} \cmidrule(lr){4-5}
$\pi^\textbf{LA}(1)$ & Share & Mean & Share & Mean\\
\midrule
FF & 11.3 & -29.2 & 1.5 & 5.0 \\
MY & 9.4 & -15.7 &  8.9 & 4.4 \\
\bottomrule
\end{tabular}}
\end{table}
Providing an aggregate overview, Table~\ref{table_distortions} presents the frequency and magnitude of probability distortions under policies $\pi_{\mathrm{FF}}^{\mathrm{LA}}(1)$ and $\pi_{\mathrm{MY}}^{\mathrm{LA}}(1)$. More specifically, we separately consider negative and positive deviations between displayed and approximated truthful probabilities $p_{j}^\pi(S_k)-\hat{\psi}_j^k$ (in both cases of at least 1 pp in absolute magnitude) and report the share of displayed probabilities affected in \% as well as the mean deviation in pp. 

Under both allocation schemes, negative distortions are more prevalent and larger (in absolute terms) than positive ones, meaning that the policies primarily steer demand by understating the likelihood of receiving the premium outcome. In case of FF, both negative and positive distortions are more aggressive in magnitude than under MY, indicating more substantial adjustments whenever demand management intervenes. By contrast, MY distorts probabilities more frequently overall (affecting 18.3\% versus 12.8\% of probabilities displayed), particularly due to more positive deviations. The remaining 87.2\% and 81.7\% of displayed probabilities closely match their approximated truthful counterparts, indicating that the performance gains achieved under $\tau=1$ relative to $\tau=0$ arise from targeted adjustments in a contained subset of decisions.

\begin{figure}
\centering 
{\includegraphics[scale = 0.5]{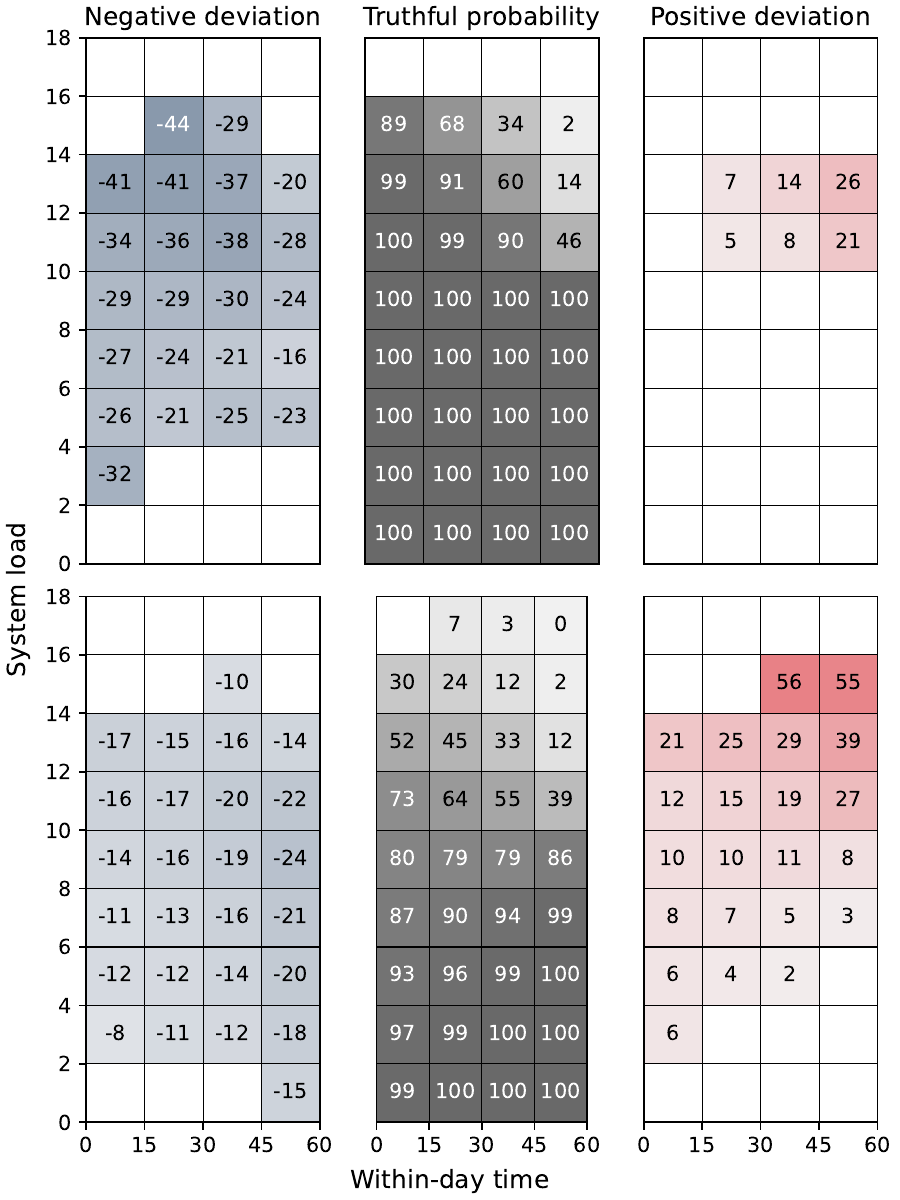}} 
\caption{State-dependent truthful probabilities and their distortion.} \label{figure_distortion}
\caption*{\footnotesize \textit{Note.} The rows of plots correspond to policies $\pi_{\mathrm{FF}}^{\mathrm{LA}}(1)$ and $\pi_{\mathrm{MY}}^{\mathrm{LA}}(1)$, respectively.} 
\end{figure}

Digging deeper, we note that both the truthful and displayed probabilities under a policy depend on the high-dimensional system state $S_k$. To gain an understanding of how probabilities get distorted across different regions of the state space, we group the observations for each pair of pre-decision state $S_k$ and hypothetical order $o_{kj}$ with $j\in\mathcal{J}^P$ encountered during the evaluation periods along two key dimensions, within-day time $t_k^w$ and system load. We define the system load as the sum of (i) the number of occupied compartments in locker $i_j^+$ and (ii) the number of pending orders for option $j$ with a higher allocation priority than $o_{kj}$. Figure~\ref{figure_distortion} presents the results in the form of heatmaps, each cell containing the average for the associated observations. The first and second rows of plots correspond to policies $\pi_{\mathrm{FF}}^{\mathrm{LA}}(1)$ and $\pi_{\mathrm{MY}}^{\mathrm{LA}}(1)$, respectively. The middle column reports the approximated truthful probabilities in \%, while the left and right columns report negative and positive deviations in pp. To ensure representativeness, we only show cells containing at least 30 observations.

The truthful probabilities are generally high when the system load remains below the total locker capacity of 10 compartments. Delivery to the locker is guaranteed under FF in these states, consistently resulting in a truthful probability of 100\%. Once the system load exceeds locker capacity, the truthful probabilities for FF decrease as the allocation decision at $t_k^w=61$ approaches: With less time remaining, occupied compartments are less likely to become available through parcel pickups before allocation.

While early arrival grants a high allocation priority under FF, it provides no comparable protection under MY. As a result, the truthful probabilities for MY may fall below 100\% even for system loads under 10. Interestingly, the within-day time has two opposing effects: On the one hand, less time until allocation reduces the likelihood for occupied compartments to become available. On the other hand, there are fewer opportunities for future orders to arrive that potentially receive a higher allocation priority. For system loads below 10, the latter effect dominates, such that the truthful probabilities under MY increase over the course of the day. At system loads above 10, the locker occupancy tends to be high, especially early in the day, and the allocation outcome of the considered order depends more strongly on capacity becoming available through parcel pickups, such that truthful probabilities decrease as the day progresses. 

Consistent with the findings from Table~\ref{table_distortions}, negative deviations are generally larger in absolute terms and more widespread than positive distortions under FF. In light of the already high truthful probabilities, there simply is little room for upward distortion. Negative distortion is strongest early in the day, when most of the demand is yet to come, and under higher system loads, when capacity is more scarce. Given the generally lower truthful probabilities in the majority of heatmap cells, negative deviations are weaker under MY and increase towards the end of the day, whereas positive deviations occur more frequently and strongly. Pronounced positive distortions towards the end of the day and under high system loads show that the provider may also steer customers by artificially inflating the attractiveness of lockers despite a substantial risk of redirection. 

\subsubsection{Redistribution of Demand} \label{cs_base_dmnd_mgmt}
To analyze how probability display and allocation decisions jointly shape which service outcomes customers receive, we randomly select one of the 30 base setting instances and examine it in more detail. We focus on all customers whose preferred service outcome (with the highest utility $u_{i}^k$) is a parcel locker $i\in\mathcal{I}^+$ and divide them into three groups: 
\begin{enumerate} 
    \item \textbf{PP} (\textit{preferred and premium}): Customers who choose the probabilistic service option containing their preferred locker and receive the locker premium outcome upon allocation. 
    \item \textbf{PB} (\textit{preferred and base}): Customers who choose the probabilistic service option containing their preferred locker but receive the associated parcel shop base outcome. 
    \item \textbf{OO} (\textit{other option}): Customers who choose a service option that does not contain their preferred locker. 
\end{enumerate}

\begin{table}
\centering
\caption{Group shares (\%).} 
{\centering
\label{table_choice_allocated}
\scriptsize
\begin{tabular}{llcccc}
\toprule
\multicolumn{2}{l}{Policy $\pi$} & $\tau$ & PP & PB & OO \\
\midrule
FF & NO & -- & 84.0 & 16.0 & 0.0 \\
 & LA & 0 & 80.4 & 6.5 & 13.1 \\
 &  & 1 & 75.0 & 2.9 & 22.0 \\ \midrule
MY & NO & -- & 84.0 & 16.0 & 0.0 \\
 & LA & 0 & 79.2 & 9.5 & 11.2 \\
 &  & 1 & 77.2 & 6.4 & 16.5 \\
\bottomrule
\end{tabular}}
\end{table}

Table~\ref{table_choice_allocated} reports the share of customers in each group relative to the total number of locker-preferring customers for selected policies. Under NO, customers choose the service option with their favorite facility by design, such that the share of group OO equals 0. Moreover, customers' choices do not differ under FF and MY, resulting in identical sets of pending orders throughout the evaluation period. Nevertheless, the allocation schemes may produce distinct locker-occupancy trajectories because pickup realizations may vary depending on which specific orders end up in the locker. In total, both schemes allocate the preferred locker to $84.0\%$ of the considered customers (PP) and redirect the remaining $16.0\%$ (PB).

The probabilistic information shared under LA and full transparency induces 13.1\% (FF) and 11.2\% (MY) of customers to forgo their preferred locker in favor of an alternative (OO). Truthful information therefore facilitates customer self-selection: Depending on the utilities associated with the preferred locker and backup parcel shop, the availability of other options, the displayed probabilities and the resulting valuations, customers proactively select alternative service options when the displayed probability for their preferred locker renders the probabilistic service option unattractive.

Removing the transparency constraints intensifies this redistribution. The share of group OO increases to $22.0\%$ ($\text{FF}$) and $16.5\%$ ($\text{MY}$). Beyond merely informing customers, probabilities displayed under $\tau=1$ act as an active demand management mechanism, guiding customers away from their preferred option (even when allocation might otherwise have been possible) to preserve capacity for other orders. FF relies on this mechanism more strongly than MY, leading to a higher share for OO and a lower share for PB.

\begin{figure}
\centering 
{\includegraphics[scale = 0.55]{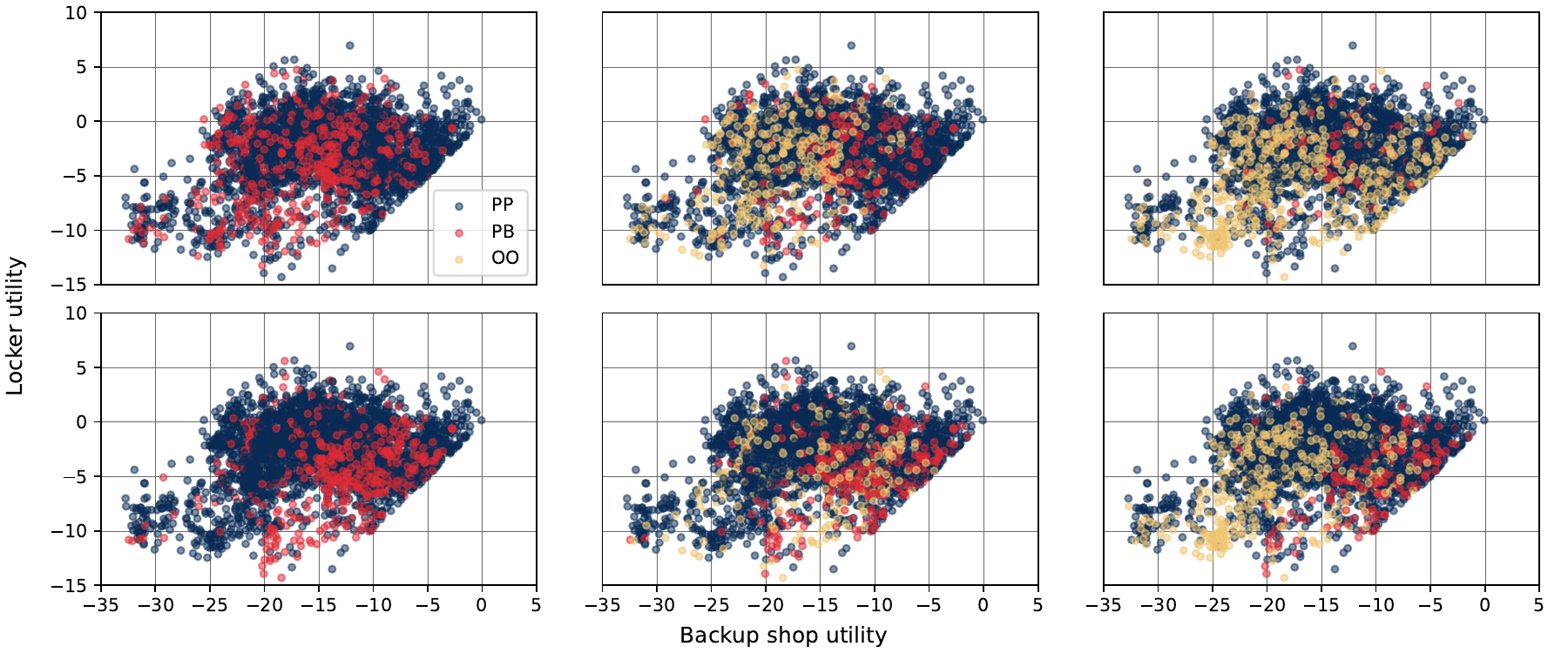}} 
\caption{Customer segmentation.} \label{figure_dmgmt}
\caption*{\footnotesize \textit{Note.} The rows of plots correspond to allocation schemes FF and MY, while the columns correspond to NO, LA under full transparency ($\tau=0$), and LA without transparency constraints ($\tau=1$).} 
\end{figure}

Figure~\ref{figure_dmgmt} provides a customer-level view of these mechanisms. In each plot, the horizontal and vertical axes represent the utilities for the backup shop and the preferred locker, respectively. Because the sample is restricted to customers whose preferred outcome is a locker, all observations lie strictly above the 45-degree diagonal, with customers positioned close to it being nearly indifferent between the two outcomes.

Under NO, both FF and MY generate identical shares for PP and PB, but Figure~\ref{figure_dmgmt} reveals that they systematically allocate the base outcome to different customers. Under FF, observations belonging to PB are distributed broadly across the utility space given that the priority for allocation solely depends on the request arrival sequence. Consequently, redirection frequently also impacts customers for whom the backup shop yields a substantial utility decrease compared to the backup shop. In contrast, MY shifts observations for PB closer toward the diagonal, targeting customers for whom the backup shop is less unattractive, which is also evident for the LA-based policies. Because customer choices under NO are independent of the allocation scheme, the improvement in customer satisfaction (Table~\ref{table_kpi_base}) from FF to MY for NO stems entirely from allocating the locker capacity to customers who value it more highly.

Under full transparency ($\tau=0$), the composition of group OO similarly depends on the allocation scheme. Under FF, this group contains many customers with relatively low backup-shop utility. Because arrival order alone determines allocation priority, these customers may face a substantial risk of receiving an unattractive base outcome and therefore avoid the probabilistic option when this risk is disclosed. Under MY, a low backup-shop utility (ceteris paribus) increases the utility loss from redirection and, consequently, the customer's allocation priority, tending to result in a high probability. Customers with low backup-shop utility are accordingly less likely to forgo their preferred-locker option under MY than under FF.

Removing the transparency constraints ($\tau = 1$) expands group OO, for FF particularly among customers exhibiting small utility differences between their preferred locker and its backup shop. Steering these nearly indifferent customers, located close to the diagonal, toward alternative options (such as the backup shop offered as a deterministic option) incurs a comparatively small utility loss. Under MY, the allocation priority criterion automatically accounts for this, resulting in less need to to steer these customers through probability distortions. The figure also indicates increased steering among customers with low backup shop utilities, especially in case of MY, which is linked to the spatial patterns discussed in the subsequent section.

\subsubsection{Customer Regret} \label{cs_base_regret}
While removing the transparency constraints boosts total customer satisfaction (Table~\ref{table_kpi_base}), the aggregate metrics do not reveal how the resulting gains and losses are distributed across individual customers. For this purpose, Table~\ref{table_regret} differentiates between customers experiencing a utility loss ($\text{CR}>0$) and those profiting from a utility gain ($\text{CR}<0$). To facilitate a direct comparison, utility gains are reported as absolute values. Averaging across evaluation streams, we report the aggregate regret (Total), the share of affected customers in percent (Share), as well as the mean, quartiles, and maximum of regret per affected customer (Mean, Q1, Q2, Q3, Max.). 

\begin{table}
\centering
\caption{Regret distribution.} 
\centering
\label{table_regret}
\scriptsize
\begin{tabular}{lcccccccccccccc}
\toprule
 & \multicolumn{7}{c}{Utility loss $(\text{CR}>0)$} & \multicolumn{7}{c}{Utility gain $(|\text{CR}<0|)$} \\ \cmidrule(lr){2-8} \cmidrule(lr){9-15}
$\pi^\textbf{LA}(1)$ & Total & Share & Mean & Q1 & Q2 & Q3 & Max. & Total & Share & Mean & Q1 & Q2 & Q3 & Max. \\ 
\midrule
FF & 1,257.2 & 12.2 & 2.5 & 0.8 & 1.7 & 3.3 & 20.4 & 2,139.5 & 7.2 & 7.3 & 3.1 & 6.6 & 10.4 & 24.4 \\
MY &  1,122.2 & 8.9 & 3.1 & 1.0 & 2.1 & 4.1 & 19.2 & 1,940.4 & 7.6 & 6.3 & 2.0 & 5.3 & 9.4 & 23.1 \\
\bottomrule
\end{tabular}
\end{table}

Under both allocation schemes, aggregate utility gains outweigh the losses, aligning with the overall increase in customer satisfaction. Notably, these outcomes are unevenly distributed: The share of customers suffering a utility loss (12.2\% for FF and 8.9\% for MY) exceeds the share of those experiencing a utility gain (7.2\% and 7.6\%), and the mean gain per affected customer is more than twice as large as the mean loss. The remaining majority of customers (80.6\% and 83.5\%) receives the identical service outcome regardless of whether transparency constraints are applied. Comparing the quartiles underscores the asymmetry in the distribution of gains and losses. Across all quartiles (Q1, Q2, Q3), the utility loss per affected customer is substantially smaller than the utility gain. Overall, a larger proportion of customers incurs a modest individual utility loss, thereby enabling substantial utility gains for a smaller group of beneficiaries.

\begin{figure}
\begin{minipage}[t]{0.55\linewidth}
\centering
{\includegraphics[scale = 0.53]{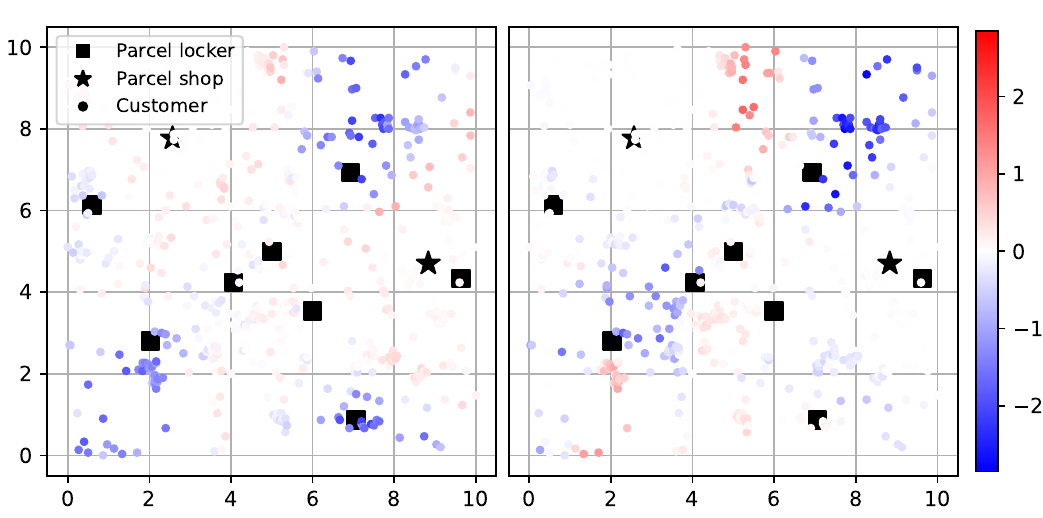}} 
\caption{Spatial regret distribution. \label{figure_spatial_regret}} 
\caption*{\footnotesize \textit{Note.} The plots represent policies $\pi_{\mathrm{FF}}^{\mathrm{LA}}(1)$ and $\pi_{\mathrm{MY}}^{\mathrm{LA}}(1)$, respectively.}
\end{minipage}
\hfill
\begin{minipage}[t]{0.45\linewidth}
\centering
{\includegraphics[scale = 0.55]{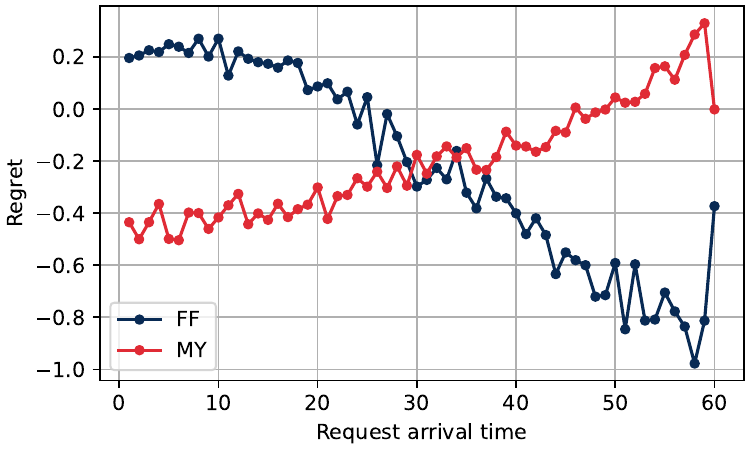}} 
\caption{Temporal regret distribution. \label{figure_temporal_regret}} 
\end{minipage}
\end{figure}

To assess whether demand management puts particular customers at a systematic (dis)advantage, Figure~\ref{figure_spatial_regret} depicts the average regret by customer location under policies $\pi_{\mathrm{FF}}^{\mathrm{LA}}(1)$ and $\pi_{\mathrm{MY}}^{\mathrm{LA}}(1)$. Complementing the spatial analysis, Figure~\ref{figure_temporal_regret} links regret to the within-day request arrival time. In both figures, positive CR values indicate a utility loss compared to full transparency, whereas negative CR implies a utility gain.

Figure~\ref{figure_spatial_regret} reveals considerable spatial heterogeneity in CR. Under FF, customers who benefit from demand management tend to be located at the periphery of the network. These customers often face long distances to alternative facilities and would incur comparatively large utility losses if denied access to their preferred locker. By steering more centrally located customers toward alternative facilities, demand management preserves scarce capacity for peripheral customers with fewer attractive substitutes. Under MY, this spatial pattern is less evident. Because MY inherently prioritizes orders according to the utility loss associated with redirection, customers located far from the backup shop already receive a high allocation priority. Consequently, there is less need to additionally protect them through demand management. Essentially, MY embeds spatial differentiation directly into the allocation decision itself, whereas FF relies more heavily on demand management to achieve a comparable effect.

The upper-right corner of the network for policy $\pi_{\mathrm{MY}}^{\mathrm{LA}}(1)$ nicely demonstrates that the spatial effects of demand management depend on the availability of alternative service options. The red cluster of customers in this region typically prefers the locker located at $(6.9,6.9)$ and faces a severe utility loss if redirected to its backup shop at $(8.8,4.7)$, which grants their orders high allocation priority under MY. At the same time, demand management can steer these customers toward another facility (such as the parcel shop at $(2.6,7.8)$ or the locker at $(5.0,5.0)$) at a comparatively small utility cost, freeing up capacity and producing utility gains for customers in the neighboring blue cluster. A second cluster of customers benefiting from demand management under MY emerges in the lower-left part of the network. These customers compete for locker capacity with peripheral customers that exhibit larger utility losses. Under full transparency, the policy tends to indiscriminately display a high probability to these peripheral customers since they have a high chance of displacing orders from customers located closer to the center. Demand management can trade off the utility loss from steering peripheral customers toward another option with the potential utility loss due to redirection, especially in case of already existing pending orders from central customers.

As a second key dimension for our regret analysis, Figure~\ref{figure_temporal_regret} illustrates systematic differences in customer regret depending on the request arrival time. Under FF and full transparency, customers arriving early possess an inherent advantage: Once an order has arrived, the provider cannot alter its ranking for allocation. To counteract this, demand management intentionally steers selected early arrivals, e.g., with comparatively low locker utility (Section~\ref{cs_base_dmnd_mgmt}), toward alternative options, at the cost of utility losses for the current customer. In anticipation of later-arriving customers, the policy preserves capacity, ultimately generating utility gains later in the day. This manifests itself in a decreasing trend for the regret curve. 

Under MY, the temporal dynamics are reversed. Later-arriving orders with larger utility losses from redirection can displace existing orders. However, even if the current customer is highly likely to receive the locker, encouraging them to select the underlying probabilistic service option by honestly communicating the probability is not necessarily optimal for aggregate customer satisfaction: Although the displacement benefits the latecomer, it can result in a greater total utility loss than if the latecomer had simply been steered toward an alternative. Demand management addresses this trade-off by guiding later-arriving customers' choices toward other options, even when their truthful probability of securing the locker is high. Consequently, probability distortion under MY tends to protect earlier orders from displacement, resulting in utility gains during the early part of the day and utility losses towards the end of the day.

In both cases, the results illustrate that demand management compensates weaknesses of the underlying allocation approaches, thereby leading to improved aggregate satisfaction. On the flip side, gains and losses are not randomly distributed across the customer population and systematically differ based on location and request arrival time.

\subsubsection{Long-Term Credibility} \label{cs_base_credibility}

\begin{figure}
\centering
{\includegraphics[scale = 0.5]{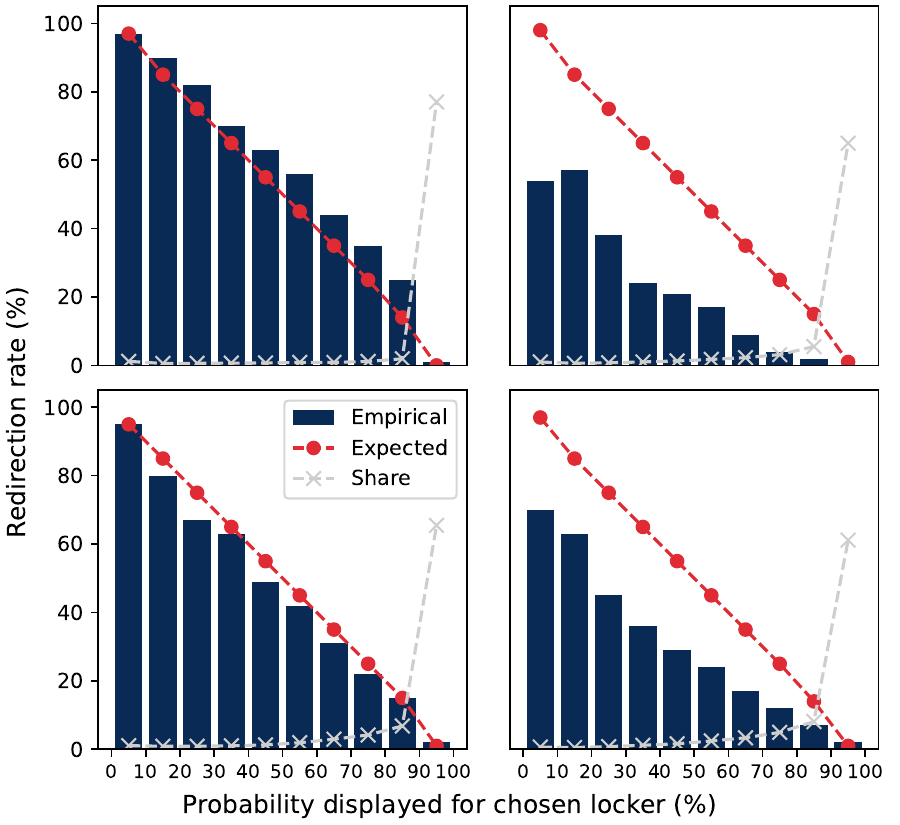}} 
\caption{Redirection rates. \label{figure_redirection_rate}} 
\caption*{\footnotesize \textit{Note.} The rows of plots correspond to allocation schemes FF and MY, while the columns correspond to LA under full transparency ($\tau=0$) and without transparency constraints ($\tau=1$).}
\end{figure}

From a managerial perspective, long-term credibility is critical; in particular when customers repeatedly interact with the system, as is the case for OOHD. To this end, Figure~\ref{figure_redirection_rate} compares the observed (empirical) redirection rates (RR) with the (expected) rates implied by the communicated probabilities. We divide the observations into bins according to the probability displayed for the option chosen by the customer and additionally report the share out of all customers per bin. The rows of plots correspond to FF and MY, whereas the columns represent LA for $\tau=0$ and $\tau=1$. 

Under full transparency, empirical and expected redirection rates mostly align, with limited discrepancies occurring in probability ranges with low customer shares. Without transparency constraints, the gap between the two rates widens substantially but continues to mostly affect sparsely populated bins. The expected RR consistently exceeds the empirical one, indicating that the policy systematically signals a higher risk of redirection than actually prevalent.

Essentially, the provider faces a trade-off between the gains from demand management and the credibility of the communicated probabilities. If customers repeatedly receive an outcome presented as unlikely, they may cease to regard the probabilities as reliable, undermining the effectiveness of probability display as a demand management lever. At the same time, customers' ability to detect systematic distortions would require frequent exposure to probability bins with strong discrepancies between expected and empirical RR.

\subsection{Impact of Setting Parameters}
Building on the base setting, we examine how variations in key problem parameters affect the value of information provision and additional gains from demand management.

\subsubsection{Systematic Utility Difference} 
To assess the role of customers' systematic preference for locker delivery, we vary the additive locker utility constant from its base value of $\beta_L=2$ to $\beta_L\in\{0,4\}$ and present the results in Table~\ref{table_kpi_beta}. Essentially, $\beta_L=0$ implies that lockers offer no systematic utility advantage over parcel shops, whereas $\beta_L=4$ raises the utility premium associated with lockers. 

\begin{table}
\centering
\caption{Performance metrics for varying locker utility constant $\beta_L$.} 
{\centering
\label{table_kpi_beta}
\scriptsize
\begin{tabular}{lllccccccccccc}
\toprule
  & & & \multicolumn{3}{c}{CS} & \multicolumn{5}{c}{(\%)} & \multicolumn{2}{c}{(pp)} \\ \cmidrule(lr){4-6} \cmidrule(lr){7-11} \cmidrule(lr){12-13}
$\beta_L$ & $\pi^\text{LA}$ & $\tau$ & Total & $\Delta$\% & pc & LS & RR & $\underline{\text{LU}}$ & $\overline{\text{LU}}$ & DC & IN & AE \\
\midrule
0 & FF & 0 & -23,076.3 & 7.6 & -5.7 & 78.6 & 4.5 & 41.7 & 94.4 & 94.8 & 0.0 & 0.6 \\
  &    & 1 & -22,449.2 & 10.1 & -5.5 & 76.0 & 2.1 & 47.9 & 90.2 & 91.7 & 17.2 & 0.3 \\ \cmidrule{2-13}
  & MY & 0 & -22,448.6 & 4.2 & -5.5 & 79.3 & 7.3 & 44.6 & 94.0 & 90.1 & 0.0 & 2.9 \\
  &    & 1 & -21,966.3 & 6.3 & -5.4 & 78.2 & 4.8 & 44.1 & 92.3 & 89.8 & 9.9 & 2.0 \\ \midrule
4 & FF & 0 & -11,356.1 & 23.3 & -2.8 & 90.4 & 5.7 & 85.6 & 97.3 & 93.0 & 0.0 & 0.7 \\
  &    & 1 & -10,375.0 & 29.9 & -2.6 & 86.2 & 3.4 & 84.8 & 93.6 & 91.0 & 21.4 & 0.4 \\ \cmidrule{2-13}
  & MY & 0 & -10,722.1 & 14.7 & -2.6 & 91.6 & 10.6 & 81.1 & 96.4 & 88.3 & 0.0 & 3.7 \\
  &    & 1 & -9,567.4  & 23.8 & -2.4 & 89.6 & 7.0 & 85.2 & 94.1 & 88.4 & 12.6 & 2.8 \\
\bottomrule
\end{tabular}}
\end{table}

In combination with Table~\ref{table_kpi_base}, the results show that customer satisfaction (CS) generally increases with $\beta_L$, reflecting the greater utility customers derive from locker delivery. More importantly, the value of providing truthful information ($\tau=0$) likewise grows with the attractiveness of lockers. Full transparency under $\beta_L=0$ improves customer satisfaction relative to NO by $7.6\%$ and $4.2\%$ for FF and MY, respectively, compared with $12.8\%$ and $7.5\%$ in the base setting. These improvements soar to $23.3\%$ and $14.7\%$ for $\beta_L=4$. 

We trace this pattern back to three mechanisms: First, a higher locker utility premium $\beta_L$ attracts more customers to lockers, as reflected in LS and LU, which intensifies competition for their limited capacity. Second, the potential utility loss from redirection to a parcel shop tends to increase because redirected customers forgo the systematic utility premium associated with lockers. Third, a higher $\beta_L$ expands the set of lockers that constitute attractive alternatives for a customer despite being located farther away than parcel shops. This gives customers more options to respond to the disclosed probabilities by switching among facilities and, consequently, increases the provider's ability to redistribute demand through information provision. Conversely, customers under $\beta_L=0$ select lockers less frequently, reducing the scarcity of locker capacity and, consequently, the value of information provision. For the same reasons, the amount of probability distortion (IN) and the additional gains in CS from deactivating the transparency constraints follow a similar pattern. Relative to full transparency, LA with $\tau=1$ improves CS by $2.7\%$ and $2.1\%$ for FF and MY under $\beta_L=0$, compared with $5.1\%$ and $4.9\%$ in the base setting as well as $8.6\%$ and $10.8\%$ for $\beta_L=4$.

\begin{figure}
\centering
{\includegraphics[scale = 0.45]{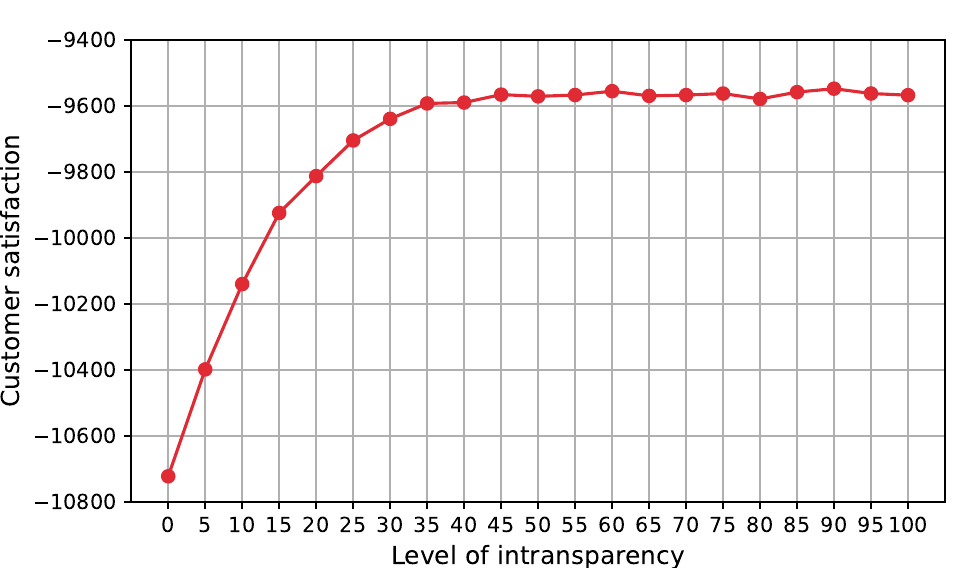}} 
\caption{Performance of policy $\pi^\text{LA}_\text{MY}(\tau)$ in setting $\beta_L=4$. \label{figure_intransparency}} 
\end{figure}

Given that we observe the largest demand management gains for $\beta_L=4$ and MY, we investigate this setting in more detail by varying $\tau$ in increments of 0.05 for policy $\pi^\text{LA}_\text{MY}(\tau)$. Figure~\ref{figure_intransparency} depicts the resulting evolution of aggregate CS. Consistent with the monotonicity established in Property~1, CS generally improves with rising $\tau$. Notably, modest relaxations of the transparency constraints already yield substantial improvements, whereas the marginal gains diminish with $\tau$. CS reaches a plateau at $\tau=0.35$, beyond which further relaxation produces no additional benefit. Overall, the results illustrate that moderate flexibility in probability displays may already suffice for effective demand management.

\subsubsection{Network Size}
\begin{table}
\centering
\caption{Performance metrics for varying network size $I$.} 
{\centering
\label{table_kpi_network}
\scriptsize
\begin{tabular}{lllccccccccccc}
\toprule
  & & & \multicolumn{3}{c}{CS} & \multicolumn{5}{c}{(\%)} & \multicolumn{2}{c}{(pp)} \\ \cmidrule(lr){4-6} \cmidrule(lr){7-11} \cmidrule(lr){12-13}
$I$ & $\pi^\text{LA}$ & $\tau$ & Total & $\Delta$\% & pc & LS & RR & $\underline{\text{LU}}$ & $\overline{\text{LU}}$ & DC & IN & AE \\
\midrule
15 & FF & 0 & -18,561.6 & 15.4 & -3.1 & 84.3 & 5.1 & 64.9 & 96.5 & 93.9 & 0.0 & 0.6 \\
   &    & 1 & -17,552.3 & 20.0 & -2.9 & 80.5 & 2.3 & 71.3 & 93.0 & 91.6 & 22.2 & 0.2 \\ \cmidrule{2-13}
   & MY & 0 & -18,000.5 & 8.8 & -3.0 & 84.3 & 8.7 & 55.7 & 95.5 & 88.7 & 0.0 & 3.4 \\
   &    & 1 & -16,867.1 & 14.6 & -2.8 & 84.2 & 6.1 & 67.1 & 94.9 & 88.4 & 12.6 & 2.3 \\ \midrule
20 & FF & 0 & -21,359.6 & 21.3 & -2.6 & 86.7 & 6.7 & 66.0 & 99.5 & 93.1 & 0.0 & 0.7 \\
   &    & 1 & -20,020.6 & 26.2 & -2.5 & 82.2 & 3.1 & 67.6 & 96.5 & 89.6 & 23.8 & 0.3 \\ \cmidrule{2-13}
   & MY & 0 & -21,007.1 & 13.1 & -2.6 & 86.9 & 10.7 & 62.7 & 99.4 & 88.4 & 0.0 & 4.6 \\
   &    & 1 & -19,023.6 & 21.3 & -2.4 & 86.3 & 7.4 & 66.0 & 98.6 & 86.6 & 15.2 & 2.8 \\ \midrule
25 & FF & 0 & -24,878.4 & 27.1 & -2.5 & 87.8 & 8.2 & 26.9 & 99.8 & 91.8 & 0.0 & 0.8\\
   &    & 1 & -22,553.7 & 33.9 & -2.2 & 82.5 & 3.2 & 35.3 & 97.1 & 87.3 & 26.2 & 0.5 \\ \cmidrule{2-13}
   & MY & 0 & -25,199.0 & 17.1 & -2.5 & 88.1 & 12.9 & 16.5 & 99.6 & 86.5 & 0.0 & 5.5\\
   &    & 1 & -21,570.1 & 29.1 & -2.1 & 87.4 & 8.5 & 32.4 & 99.3 & 83.4 & 17.6 & 3.2 \\
\bottomrule
\end{tabular}}
\end{table}

To test the scalability of our method, we increase the number of facilities $I$ in the network, thereby modifying and expanding the set of offered service options. Because the total service area remains fixed, this effectively results in a denser network with shorter distances between customers and their closest facilities. Table~\ref{table_kpi_network} demonstrates that the mean satisfaction per customer (pc) improves in response. Note that we scale demand proportionally with $I$, leading to a decreasing value for total CS (despite the improvement in mean utility per customer) because the provider serves a larger number of customers. 

The value of truthful information provision increases with network size. Relative to NO, full transparency improves CS under FF by $15.4\%$, $21.3\%$, and $27.1\%$ for $I\in\{15,20,25\}$, respectively. Under MY, the corresponding improvements climb from $8.8\%$ to $13.1\%$ and $17.1\%$. As the network becomes denser, customers choose from a broader set of attractive facilities, and truthful probabilities become increasingly useful for differentiating among them. Notably, RR also rises with the network size under both allocation schemes. In a denser network, the greater number of parcel shops reduces the likelihood that the base outcome associated with a locker is located far from the customer, making redirection less inconvenient. Consequently, customers become more willing to accept this risk. These same aspects also lead to denser OOHD networks offering more potential for demand management. Relative to full transparency, removing the constraints improves CS under FF by $5.4\%$, $6.3\%$, and $9.3\%$ for $I\in\{15,20,25\}$, respectively, with the corresponding gains ranging from $6.3\%$ to $9.4\%$ and $14.4\%$ under MY.

\section{Conclusion and Outlook} \label{conclusion}
In this work, we introduce probabilistic service options to dynamically communicate and manage uncertainty about future service outcomes. Upon the arrival of a request, the provider displays the probabilities with which the customer will receive a capacity-constrained premium outcome at the time of service provision. Unlike existing work on probabilistic products, which treats the probabilities as a static design choice, the truthful probabilities in our setting emerge endogenously from the interaction of future demand, customer choices, allocation decisions, and uncertain capacity utilization, effectively depending both on the system state and the policy applied by the provider. By demonstrating that probability displays can serve as a novel information-based demand management lever, our work raises a fundamental question regarding the extent to which providers should be permitted to distort probabilities to optimize aggregate system performance.

As novel modeling contributions, we formulate a recursive equation to compute the truthful probabilities and propose transparency constraints that limit the deviations between displayed and truthful probabilities. The underlying idea of recursively evaluating the state-dependent probability that a certain event will occur under a policy is not limited to our problem and can be transferred to any sequential decision process. Because the exact computation is intractable for realistic instances, we develop an adapted reinforcement learning procedure as an approximation. We are confident that our method of constructing policies based on approximated, policy-dependent probabilities of future events is a promising starting point for future work.

Managerially, our results indicate that dynamic information provision is a valuable tool to counteract imbalances in demand. Additionally, we find that targeted adjustments of probabilities and a slight relaxation of the allowed level of intransparency can already generate considerable improvements. However, certain customer groups may be put at a systematic disadvantage, and the provider must trade off the gains attainable through demand management with the potential erosion of credibility if distortions are detected.

To conclude, we highlight two compelling directions for future research:
\begin{enumerate}
    \item \textit{Advanced allocation mechanisms.} We restrict attention to non-anticipatory and pre-determined allocation schemes, which enables us to establish the uniqueness of fully transparent probability displays and to isolate the effects of demand management. Future research could study anticipatory allocation strategies where the provider takes the future evolution of the system into account when allocating service outcomes. Another important extension would be the joint optimization of probability display and allocation and the potential feedback loops between the two decision types.
    \item \textit{Customer learning.} We assume that customers do not update their beliefs about the accuracy of the communicated probabilities and the provider's trustworthiness over time. Incorporating such dynamics in future work would uncover whether demand management gains persist under this long-term perspective.
\end{enumerate}
\newpage

\begin{appendices}
\section{Notation} \label{overview_notation} 

{\footnotesize
\renewcommand{\arraystretch}{1.5}
\begin{longtable}{p{5.3cm} p{10.5cm}}
$i\in\mathcal{I}=\mathcal{I}^-\cup\mathcal{I}^+$        &   Service outcomes, partitioned into base outcomes $\mathcal{I}^-$ and premium outcomes $\mathcal{I}^+$ \\
$Q_i$                                                   &   Capacity of premium outcome $i\in\mathcal{I}^+$ \\
$j\in\mathcal{J}=\mathcal{J}^P\cup\mathcal{J}^D$        &   Service options, partitioned into probabilistic service options $\mathcal{J}^P$ and deterministic service options $\mathcal{J}^D$ \\
$(i_j^-,i_j^+)$                                         &   Base outcome $i_j^-\in\mathcal{I}^-$ and premium outcome $i_j^+\in\mathcal{I}^+$ of probabilistic service option $j\in\mathcal{J}^P$ \\
$i_j$                                                   &   Service outcome $i_j\in\mathcal{I}^-$ of deterministic service option $j\in\mathcal{J}^D$ \\
$t\in\mathcal{T}$                                       &   Point in time \\
$p_j$                                                   &   Displayed probability for premium outcome of probabilistic service option $j\in\mathcal{J}^P$ \\
$t\in\mathcal{T}^A\subset\mathcal{T}$                   &   Points in time for allocation \\
$k=1,\dots,\infty$                                      &   Decision epoch \\
$S_k=(t_k,r_k,\mathcal{O}_k,\mathcal{C}_k)\in\mathcal{S}$ & Pre-decision state in decision epoch $k$ \\
$t_k$                                                   &   Point in time in decision epoch $k$ \\
$r_k$                                                   &   Request in decision epoch $k$ \\
$t^A_{r_k}\in\mathcal{T}^A$                             &   Allocation point in time for request in decision epoch $k$ \\
$o\in\mathcal{O}_k$                                     &   Pending orders in decision epoch $k$ \\
$j_o\in\mathcal{J}$                                     &   Service option chosen by order $o$ \\
$t^A_{o}\in\mathcal{T}^A$                               &   Allocation point in time for order $o$ \\
$\mathcal{C}_k=\{\mathcal{C}_{ik}:i\in\mathcal{I}^+\}$  &   Capacity utilization in decision epoch $k$ \\
$q_i$                                                   &   Available capacity of premium outcome $i\in\mathcal{I}^+$ \\
$S_0$                                                   &   Initial state \\
$X_k\in\mathcal{X}(S_k)$                                &   Decision in decision epoch $k$ \\
$P_k=(p_j^k)_{j\in\mathcal{J}^P}$                       &   Probability display decision in decision epoch $k$ \\
$\mathcal{O}_k^A=\{o\in\mathcal{O}_k:t_o^A=t_k\}$                 &   Orders to be allocated in decision epoch $k$ \\
$A_k=(a_o^k)_{o\in\mathcal{O}_k^A}$                     &   Allocation decision in decision epoch $k$ \\
$W_{k+1}=(j_{r_k},\mathcal{C}_{k+1},r_{k+1},t_{k+1})\in\mathcal{W}$ & Exogenous information revealed during transition from decision epoch $k$ to $k+1$ \\
$\mathbb{P}(W_{k+1}|S_k,X_k)$                           &   Probability distribution of exogenous information \\
$u_i^k$                                                 &   Utility that request in decision epoch $k$ associates with service outcome $i\in\mathcal{I}$ \\
$U_j^k=U(P_k,u_{i_j^-},u_{i_j^+})$                      &   Valuation for service option $j\in\mathcal{J}$ \\
$j_{r_k}\in\mathcal{J}$                                 &   Service option chosen by the request in decision epoch $k$ \\
$R(S_k,X_k,W_{k+1})$                                    &   Reward function \\
$\pi\in\Pi$                                             &   Policy \\
$X^\pi_k(S_k)$                                          &   Decision in pre-decision state $S_k$ under policy $\pi$ \\
$\pi^*\in\Pi$                                           &   Optimal policy \\
$\bar{R}^\pi$                                           &   Expected average reward per decision epoch under policy $\pi$ \\
$S_k^c=(t_k,r_k,\mathcal{O}_{k+1},\mathcal{C}_k,\mathcal{O}_k^A,A_k)\in\mathcal{S}^c$ & Post-choice state in decision epoch $k$ \\
$W_{k+1}^c=j_{r_k}\in\mathcal{W}^c$                     &   Exogenous information for transition from pre-decision state to post-choice state \\
$S_k^c=S^{MW^c}(S_k,X_k,W_{k+1}^c)$                     &   Transition function from pre-decision state to post-choice state \\
$W_{k+1}^s=(\mathcal{C}_{k+1},r_{k+1},t_{k+1})\in\mathcal{W}^s$ & Exogenous information for transition from post-choice state to pre-decision state \\
$S_{k+1}=S^{MW^s}(S_k^c,W_{k+1}^s)$                     &   Transition function from post-choice state to pre-decision state \\
$\psi_o^\pi(S_k^c)$                                     &   Truthful probability \\
$\mathbb{P}(W_{k+1}^c|S_k,X_k),\mathbb{P}(W_{k+1}^s|S_k^c)$ & Probability distribution of exogenous information \\
$V(S)$                                                  &   Value function in pre-decision state $S\in\mathcal{S}$ \\
$V^c(S)$                                                &   Value function in post-choice state $S\in\mathcal{S}^c$ \\
$R^c(S_k,X_k,W_{k+1}^c)$                                &   Reward function for transition from pre-decision state $S_k$ to post-choice state $S_k^c$ \\
$R^s(S_{k}^c,W_{k+1}^s)$                                &   Reward function for transition post-choice state $S_k^c$ to pre-decision state~$S_{k+1}$ \\
$\mathcal{S}^p=\{S_k\in\mathcal{S}:r_k\neq 0\}$         &   Set of pre-decision states with probability display decision \\
$P^\pi(S_k)=(p_j^\pi(S_k))_{j\in\mathcal{J}^P}$         &   Probabilities displayed under policy $\pi$ in pre-decision state $S_k$ \\
$o_{kj}$                                                &   Hypothetical order for probabilistic service option $j\in\mathcal{J}^P$ in pre-decision state $S_k$ \\
$S_{kj}^c=S^{MW^c}(S_k,P^\pi(S_k),j)$                   &   Hypothetical post-choice state for probabilistic service option $j\in\mathcal{J}^P$ and pre-decision state $S_k$ \\
$\tau\in[0,1]$                                          &   Level of intransparency \\
$\Pi(\tau)$                                             &   Set of feasible policies for level of intransparency $\tau$ \\
$\alpha$                                                &   Allocation scheme \\
$\hat{\psi}_o(S_k^k)$                                   &   Approximated truthful probability \\
$\hat{\psi}_{\bm{\theta}}$                                   &   Neural network \\
${\bm{\theta}}$                                              &   Parameter weights of neural network \\
${\bm{\phi}}(S_k^c,o)$                                       &   Feature vector for neural network \\
$e\in\mathcal{E}$                                       &   Experience in replay memory \\
$n=1,\dots,N$                                           &   Decision epoch simulated for offline training \\
$\hat{\psi}_j^n=\hat{\psi}_{\bm{\theta}}({\bm{\phi}}(S_{nj}^c,o_{nj}))$ & Approximated truthful probability for probabilistic service option $j\in\mathcal{J}^P$ in decision epoch $n$ \\
$(o_e,{\bm{\phi}}_e,a_e$)                                    &   Order, feature vector, and allocation outcome associated with experience~$e$ \\
$\mathcal{B}\subset\mathcal{E}$                         &   Sampled mini-batch of experiences \\
$\mathcal{L}({\bm{\theta}},\mathcal{B})$                     &   Binary cross-entropy loss \\
$f=1,\dots,F$                                           &   Sampled trajectories \\
$a_{of}$                                                &   Allocation outcome for order $o$ in trajectory $f$ \\
$\hat{\psi}^{MC}_o$                                     &   Monte Carlo estimate for truthful probability \\
$\hat{V}^c(S_k^c)$                                      &   Approximated post-choice state value function \\
$\omega\in\Omega$                                       &   Utility scenario \\
$\Tilde{u}_{i\omega}$                                   &   Sampled utility for service outcome $i\in\mathcal{I}$ in utility scenario $\omega$ \\
$\Tilde{U}_{j\omega}$                                   &   Valuation for service option $j\in\mathcal{J}$ in utility scenario $\omega$ \\
$\Tilde{U}_\omega$                                      &   Maximum valuation in utility scenario $\omega$ \\
$c_{j\omega}$                                           &   Customer choice in utility scenario $\omega$  \\
$\hat{\mathbb{P}}(W_{k+1}^c=j|S_k,X_k)$                 &   Approximated choice probability \\
$M,M',M_\omega$                                         &   Sufficiently large constants \\
$t_k^d=1,\dots,\infty$                                  &   Day in decision epoch $k$ \\
$t_k^w=1,\dots,T+1$                                     &   Within-day time in decision epoch $k$ \\
$v_i^k$                                                 &   Deterministic utility that request in decision epoch $k$ associates with service outcome $i\in\mathcal{I}$ \\
$\beta_L$                                               &   Additive locker utility constant \\
$\delta_{r_k i}$                                        &   Euclidean distance between location of customer $r_k$ and facility $i\in\mathcal{I}$ \\
$\beta_D$                                               &   Distance utility coefficient \\
$\varepsilon_i^k$                                       &   Stochastic utility that request in decision epoch $k$ associates with service outcome $i\in\mathcal{I}$ \\
$\hat{u}_o^-,\hat{u}_o^+$                               &   Conditional expected utility for base and premium outcome of order $o$ \\
$I=|\mathcal{I}|$                                       &   Number of facilities \\
$\hat{V}^c(S_{kj},\omega)$                              &   Approximated post-choice state value in utility scenario $\omega$ \\
$\pi_\alpha$                                            &   Policy for given allocation scheme $\alpha$ \\
$h_o$                                                   &   Time horizon of order $o$ \\
$\mathcal{U}(h)$                                        &   Induction statement \\
$l\in\mathcal{L}$                                       &   Lookahead scenario \\
$\hat{V}_l^-$                                           &   Approximated post-choice state value without hypothetical order $o_{kj}$ in lookahead scenario $l$ \\
$\hat{V}_{j\omega l}^+$                                 &   Approximated post-choice state value with hypothetical order $o_{kj}$ in utility scenario $\omega$ and lookahead scenario $l$ \\
$\Delta_{j\omega l}$                                    &   Incremental value of service option $j\in\mathcal{J}$ in utility scenario $\omega$ and lookahead scenario $l$ \\
$\Delta_{j\omega}$                                      &   Average incremental value of service option $j\in\mathcal{J}$ in utility scenario $\omega$ \\
$\hat{V}^-$                                             &   Average approximated post-choice state value without hypothetical order~$o_{kj}$ \\
$\hat{V}_{j\omega}^+$                                   &   Average approximated post-choice state value with hypothetical order $o_{kj}$ in utility scenario $\omega$ \\
$\mathcal{P}(S_k)$                                      &   MIP solution space \\
$d^*_{jl}$                                         &   Displacement due to hypothetical order $o_{kj}$ in lookahead scenario $l$ \\
\end{longtable}}

\section{Uniqueness of Fully Transparent Probability Displays Under Non-Anticipatory Allocation} \label{proofs}
Before establishing the property, we note that non-anticipatory allocation induces a one-directional dependence across allocation times $t\in\mathcal{T}^A$: A pending order can affect the allocation outcome of another order only if it is allocated earlier or as part of the same allocation decision. Conversely, pending orders allocated strictly later do not affect earlier allocation decisions. Moreover, allocation generally depends on the realized customer choices, but not directly on the probabilities previously displayed to individual customers.\\[-0.25cm]

\textbf{Property:} \textit{Consider a fixed, fully specified, and non-anticipatory allocation scheme $\alpha$. For every pre-decision state $S_k\in\mathcal{S}:t_k\in\mathcal{T}^A$, the allocation scheme $\alpha$ uniquely determines the joint allocation outcome for all orders $o\in\mathcal{O}_k^A$ independent of pending orders $o'\in\mathcal{O}_k:t_{o'}^A>t_k$. Then, conditional on $\alpha$, the truthful probabilities and thus the probabilities displayed under full transparency $\tau=0$ are unique for every pre-decision state $S\in\mathcal{S}^p$. Therefore, it suffices to identify the single policy feasible under the transparency constraints to find the optimal policy $\pi^*_\alpha$ given $\alpha$.}\\[-0.25cm]

\textit{Proof.}
For an order $o$ arising at decision epoch $k$, define its time horizon in the system by $h_o=t^A_o-t_k$. Because (i) the points in time $t_k$ are discrete, (ii) $t^A_o$ is finite, and (iii) probability display and allocation occur at distinct decision epochs, it holds that $h_o\in\mathbb{N}$.

For every $h\in\mathbb{N}$, let $\mathcal{U}(h)$ denote the following statement: \textit{For every probability display pre-decision state $S_k\in\mathcal{S}^p$ and hypothetical order $o_{kj}$ arising if request $r_k$ chooses service option $j\in\mathcal{J}^P$ with $h_{o_{kj}}=h$, the truthful allocation probabilities $\psi^\pi_{o_{kj}}(S_{kj}^c)$ for $o_{kj}$ exist and are uniquely determined. Consequently, under full transparency, the probability-display vector $P^{\pi_\alpha}(S_k)$ exists and is unique.} Note that the allocation time $t_{r_k}^A$ is an attribute of the request such that all service options $j\in\mathcal{J}$ lead to the same horizon $h_{o_{kj}}$. We prove $\mathcal{U}(h)$ for every $h\in\mathbb{N}$ by induction.

For the base case, consider $h=1$, meaning that the hypothetical order $o_{kj}$ arising in $k$ is allocated at the subsequent decision epoch $k+1$. By the prescribed ordering of events, no additional probability display decision occurs between the creation and allocation of $o_{kj}$. Hence, every order that can affect its allocation is already represented in post-choice state $S_{kj}^c$, and the exogenous-information distribution $\mathbb{P}(W_{k+1}^s|S_{kj}^c)$ together with the transition function $S_{k+1}=S^{MW^s}(S_{kj}^c,W^s_{k+1})$ uniquely determine the distribution of the next pre-decision state $S_{k+1}$. Conditional on each realization of the latter, the fully specified allocation scheme $\alpha$ determines a unique joint allocation outcome for all orders $o\in\mathcal{O}_{k+1}^A$, which includes $o_{kj}$. Therefore, the truthful allocation probabilities of $o_{kj}$ for every $j\in\mathcal{J}^P$ exist and are unique. Under full transparency, the displayed probabilities must coincide with the truthful probabilities, establishing $\mathcal{U}(1)$.

For the induction step, fix an arbitrary $h\geq2$ and suppose that $\mathcal{U}(h')$ holds for every $h'<h$. Consider an arbitrary pre-decision state $S_k\in\mathcal{S}^p$ with hypothetical order $o_{kj}$ satisfying $t^A_{o_{kj}}-t_k=h$. Additional requests may arise before $t_{o_{kj}}^A$ and generate orders allocated before or jointly with $o_{kj}$. Consider any such order $o'$ arising at a later decision epoch $k'$ with $t_k<t_{k'}<t^A_{o_{kj}}$. There are two relevant cases to distinguish.

\begin{itemize}
    \item Case 1: $t^A_{o'}\leq t^A_{o_{kj}}$. Because $\alpha$ is non-anticipatory, $o'$ can affect the allocation outcome of $o_{kj}$ only if $t^A_{o'}\leq t^A_{o_{kj}}$. Its horizon $h_{o'}$ satisfies
    \[
    \begin{aligned}
    h_{o'} =
    t^A_{o'}-t_{k'}\leq t^A_{o_{kj}}-t_{k'}< t^A_{o_{kj}}-t_k=h
    \end{aligned}
    \]
    Since probability display and allocation occur at distinct decision epochs, $1\leq h_{o'}<h$. Hence, every future order capable of affecting the allocation of $o_{kj}$ has a horizon in $\{1,\ldots,h-1\}$ and is covered by the induction hypothesis. The truthful allocation probabilities, and therefore the fully transparent display vector at epoch $k'$, are thus uniquely determined. Since the customer-choice model $\mathbb{P}(W_{k'+1}^c|S_{k'},P^{\pi_\alpha}(S_{k'}))$ uniquely maps the state and probability display vector into a choice probability distribution, the conditional distribution of the customer's choice is unique.
    \item Case 2: $t^A_{o'}> t^A_{o_{kj}}$. The order $o'$ is allocated strictly after order $o_{kj}$. Following its creation, $o'$ is included in the updated set of pending orders $O_{k'+1}$, but it does not occupy capacity before its allocation. Moreover, by non-anticipativity of $\alpha$, neither the existence nor the characteristics of $o'$ affect any allocation decision made at or before $t^A_{o_{kj}}$. Additionally, the presence of $o'$ does not influence the conditional distribution of the exogenous information relevant to the allocation of $o_{kj}$, including capacity availability, request arrivals, and customer choices that may affect allocations up to $t^A_{o_{kj}}$. Consequently, conditional on any realization of the process up to $t^A_{o_{kj}}$, the allocation outcome of $o_{kj}$ is invariant to $o'$. Hence, when constructing the conditional distribution of the allocation state for $o_{kj}$, the customer choice and subsequent state information associated exclusively with $o'$ can be marginalized out. Orders allocated strictly after $o_{kj}$ therefore do not affect the truthful allocation probability of $o_{kj}$.
\end{itemize}

The finite composition of state transitions, which follow unique conditional distributions, together uniquely determines the probability distribution of the allocation state for order $o_{kj}$ and the associated set of orders to be allocated. Conditional on each realization, scheme $\alpha$ uniquely determines the joint allocation outcome or its probability distribution. Hence, the truthful allocation probability $\psi_{o_{kj}}^{\pi_\alpha}(S_{kj}^c)$ exists and has a unique value. Therefore, the probability display vector in pre-decision state $S_k$ is unique, and $\mathcal{U}(h)$ holds.

By induction, $\mathcal{U}(h)$ holds for every $h\in\mathbb{N}$. Since $S_k$ and $j$ were arbitrary, the truthful allocation probabilities exist and are uniquely determined in every post-choice state. Consequently, under $\tau=0$, the probability display decision is unique in every pre-decision state. Therefore, the full-transparency constraints admit a unique policy conditional on $\alpha$, which is the optimal policy $\pi_\alpha^*$ given $\alpha$.

\section{Policies} \label{policy_details}
We provide details on our lookahead-model-based approximation for the post-choice state values, the features for predicting truthful probabilities, and the hyperparameters used in our computational study.

\subsection{Lookahead Model}
With our lookahead model, we approximate the post-choice state values $\hat{V}^c(S^c_{kj})$ in (\ref{MIP_objective}) using a set of sampled lookahead scenarios $l\in\mathcal{L}$ that represent the possible future evolution of the system. More precisely, the lookahead scenarios extend until the end of the current day $(t_k^d,T+1)$ and comprise sampled pickup times for parcels currently occupying locker capacity, drawn from their conditional distributions, as well as future requests and their utilities. In the following, we omit the dependence on $S^c_{kj}$ for notational convenience.

As a reference, we first compute the value $\hat{V}_{l}^{-}$ of the system without hypothetical order $o_{kj}$ in each lookahead scenario $l$. For each service option $j\in\mathcal{J}$ and utility scenario $\omega\in\Omega$, we then compare it to the corresponding value $\hat{V}_{j\omega l}^+$ if $o_{kj}$ is included. This value depends on $\omega$ because it contains the current request's utility $\Tilde{u}_{i_j^-\omega}$ or $\Tilde{u}_{i_j^+\omega}$ sampled for the allocated service outcome (Section~\ref{s_aopt}). To reduce the computational burden, we estimate $\hat{V}_l^-$ once per lookahead scenario $l$ and then heuristically determine $\hat{V}_{j\omega l}^+$ through incremental adaptations. 

For the baseline $\hat{V}_l^-$, we formulate a full-information assignment problem without the hypothetical order. Specifically, the objective coefficients correspond to the conditional expected utilities of currently pending orders and the sampled utilities of future orders, while the sampled pickup times determine the available locker capacity. We prevent infeasible assignments of existing pending orders to service outcomes not contained in their chosen service options by setting the objective coefficients to $-\infty$. We solve the assignment problem to optimality with a modified Jonker-Volgenant algorithm \citep{crouse_2016}, implemented in SciPy~1.16.

We define the incremental value of service option $j$ in utility scenario $\omega$ and lookahead scenario $l$ as $\Delta_{j\omega l}=\hat{V}_{j\omega l}^+-\hat{V}_l^-$. Thus, $\Delta_{j\omega l}$ captures the additional utility generated by the hypothetical order as well as the potential displacement of orders. Averaging the incremental value for service option $j$ and utility scenario $\omega$ across lookahead scenarios yields $\Delta_{j\omega}=\frac{1}{|\mathcal{L}|}\sum_{l\in\mathcal{L}}\Delta_{j\omega l}$. With the average baseline value $\hat{V}^-=\frac{1}{|\mathcal{L}|}\sum_{l\in\mathcal{L}}\hat{V}_l^-$, we obtain $\hat{V}_{j\omega}^+=\hat{V}^-+\Delta_{j\omega}$. Recall that the immediate reward for probability display in our OOHD setting equals $R(S_k,P_k,j)=0$. Substituting $\hat{V}_{j\omega}^+=\hat{V}^-+\Delta_{j\omega}$ into the MIP objective~(\ref{MIP_objective}) and using $\sum_{j\in\mathcal{J}}c_{j\omega}=1\ \forall\omega\in\Omega$ yields \begin{align} \sum_{j\in\mathcal{J}}\frac{1}{|\Omega|}\sum_{\omega\in\Omega}c_{j\omega}\hat{V}_{j\omega}^+=\hat{V}^-+\sum_{j\in\mathcal{J}}\frac{1}{|\Omega|}\sum_{\omega\in\Omega}c_{j\omega}\Delta_{j\omega} \label{MIP_inc_obj}
\end{align}
Because $\hat{V}^-$ is independent of the displayed probabilities $P_k$, it can be omitted from (\ref{MIP_inc_obj}). Letting $\mathcal{P}(S_k)$ denote the solution space formalized by Constraints~(\ref{MIP_compute_valuations})-(\ref{MIP_domains}), we obtain the incremental-value-based formulation for the MIP: 
\begin{align} 
    \max_{P_k\in\mathcal{P}(S_k)}\left\{\frac{1}{|\Omega|}\sum_{\omega\in\Omega}\sum_{j\in\mathcal{J}}c_{j\omega}\Delta_{j\omega}\right\} \label{incremental_value_objective}
\end{align}

Starting from the optimal baseline solution computed for $\hat{V}_l^-$, we heuristically derive the incremental value $\Delta_{j\omega l}$ based on the change in the objective value caused by including hypothetical order $o_{kj}$ and allowing up to one reassignment. Currently pending orders $o\in\mathcal{O}_k$ may be reassigned only to the base outcomes associated with their chosen service options, whereas sampled future orders may be reassigned to any facility with sufficient capacity.

If the hypothetical order corresponds to a probabilistic service option $j\in\mathcal{J}^P$, three cases may arise: 
\begin{enumerate} 
    \item If locker $i_j^+$ has available capacity in lookahead scenario $l$, the hypothetical order receives the premium outcome without displacing another order, and the incremental value equals $\Delta_{j\omega l}=\widetilde{u}_{i_j^+\omega}$. 
    \item If locker $i_j^+$ is fully occupied in the baseline solution for $l$ but sufficient capacity can be released by reassigning another order, we identify the feasible reassignment that causes the smallest loss in the objective value. Let $d_{jl}^*\leq0$ denote the corresponding objective-value change. The hypothetical order can either receive the premium outcome $i_j^+$ at the cost of $d_{jl}^*$ due to the reassignment or the base outcome $i_j^-$, depending on which alternative results in a higher value: $\Delta_{j\omega l}=\max\left\{\widetilde{u}_{i_j^+\omega}+d_{jl}^*,\widetilde{u}_{i_j^-\omega}\right\}$.
    \item If locker $i_j^+$ is full and no feasible reassignment exists, the hypothetical order receives the base outcome $i_j^-$, leading to an incremental value of $\Delta_{j\omega l}=\widetilde{u}_{i_j^-\omega}$.
\end{enumerate} 
In case of a deterministic option $j\in\mathcal{J}^D$, the hypothetical order receives the associated outcome, yielding $\Delta_{j\omega l}=\widetilde{u}_{i_j\omega}$.

Finally, we take the average across lookahead scenarios to obtain $\Delta_{j\omega}=\frac{1}{|\mathcal{L}|}\sum_{l\in\mathcal{L}}\Delta_{j\omega l}$ and solve the incremental-value-based MIP~(\ref{incremental_value_objective}). Without this incremental strategy, we would have to solve a separate assignment problem for every combination of service option, utility scenario, and lookahead scenario, which conflicts with the requirement that customers expect to view the displayed probabilities in real time.

\subsection{Features}
\begin{table}
\centering
\caption{Feature overview.} 
{\renewcommand{\arraystretch}{1.2}
\centering
\label{features}
\scriptsize
\begin{tabular}{m{1cm} m{1cm} m{1cm} m{11cm}} \hline
Feature & Min. & Max. & Interpretation \\
\hline
$t^A$ & $1$ & $T$ & Number of within-day time periods until allocation $T+1-t_k$\\
$O^+$ & $0$ & $T$ & Number of pending orders for service option $j$ with a higher priority for allocation than order $o$\\
$C^d$ & $0$ & $Q_{i_j^+}$ & Number of occupied capacity units in locker $i_j^+$ with elapsed storage duration of $d$ days \\
$x^l,y^l$ & $\underline{x}, \underline{y}$ & $\bar{x},\bar{y}$ & Coordinates of locker $i_j^+$\\
$s$ & $\underline{s_i}$ & $\overline{s_i}$ & Expected demand share of locker $i_j^+$\\
$\Delta v$ & $\underline{\Delta v}$ & $\overline{\Delta v}$ & Deterministic utility component difference $v_{i_j^+}-v_{i_j^-}$ for order $o$\\
$x^c,y^c$ & $\underline{x^c}, \underline{y^c}$ & $\bar{x^c},\bar{y^c}$ & Coordinates of customer location\\
$\Tilde{q}^B$ & 0 & $Q_{i_j^+}$ & Expected available capacity in locker $i_j^+$ at the time of allocation $(t_k^d,T+1)$ \\
$\hat{\psi}^B$ & 0 & 1 & Baseline probability estimate \\
\hline
\end{tabular}}
\end{table}
Table~\ref{features} summarizes our features ${\bm{\phi}}(S_{k}^c,o)$ for approximating the truthful probabilities $\hat{\psi}_{\bm{\theta}}$. Beyond the information directly available from the state variables and global instance parameters, we estimate the expected demand share $s_i$ for each locker $i\in\mathcal{I}^+$ ex ante by simulating 100 days with probability display NO (Section~\ref{cs_policies}) and computing the percentage of customers choosing the corresponding service option. To construct the lookahead-based features $\Tilde{q}^B$ and $\hat{\psi}^B$, we use the same simulated order streams to generate pools of future orders for each within-day period $t_k^w$. These pools are independent of the remaining state information and can therefore be generated and stored in advance. For a given post-choice state and order $o$, we use the corresponding order pool to estimate the average number of future arriving orders with higher allocation priority than $o$, yielding a baseline estimate for the probability that $o$ receives the premium outcome. Additionally, we track the locker capacity expected to be available at the time of allocation.

\subsection{Hyperparameters}
We use the following hyperparameters across all experimental settings:

To obtain the approximation $\hat{\psi}_{\bm{\theta}}$, we simulate 5000 days of training. Starting from an initial value of 1, the probability for exploration decays exponentially to 0.05 over the first 4000 days and remains constant afterward. The neural network is implemented using PyTorch 2.5.1 and comprises 2 hidden layers with 32 neurons each. The hidden layers use rectified linear unit (ReLU) activation functions, while the output layer employs a sigmoid activation function. The replay memory $\mathcal{E}$ stores experiences for all pending orders corresponding to the 250 most recently encountered post-choice states. For each network update, we randomly sample 4 post-choice states represented in $\mathcal{E}$ and group all the associated experiences for the individual pending orders into mini-batches with at most 64 observations. We update the network parameters ${\bm{\theta}}$ using the AdamW optimizer \citep{loshchilov&hutter_2019} with a learning rate of 0.0005 and the gradient norm clipped to 1.

For probability display decisions, the MIP is based on $|\Omega|=50$ utility scenarios. We replace the sufficiently large constant $M$ with a scenario-specific value $M_\omega=\max_{i\in\mathcal{I}} \{\Tilde{u}_{i\omega}\}-\min_{i\in\mathcal{I}} \{\Tilde{u}_{i\omega}\}$ and set $M'=\min\{0.001,\hat{\psi}_1^k,\dots,\hat{\psi}_{|\mathcal{J}|}^k\}$. For LA, we approximate the post-choice state values with $|\mathcal{L}|=10$ lookahead scenarios.

\end{appendices}

\bibliographystyle{apalike}
\bibliography{references}

\end{document}